\documentclass[11pt]{article}

\usepackage[preprint]{acl}

\usepackage{times}
\usepackage{latexsym}

\usepackage[T1]{fontenc}

\usepackage[utf8]{inputenc}

\usepackage{microtype}

\usepackage{inconsolata}

\usepackage{graphicx}
\usepackage{booktabs}
\usepackage{multirow}

\usepackage{kotex}
\usepackage{amsfonts}
\usepackage{amssymb}
\usepackage{amsmath}

\usepackage{algorithm}
\usepackage{algpseudocode}
\usepackage{algorithmicx}

\usepackage[disable]{todonotes}
\usepackage{subcaption}
\usepackage{placeins}

\usepackage{fvextra}

\title{TLPO: Token-Level Policy Optimization for Mitigating Language Confusion in Large Language Models}

\author{
  \textbf{Jinho Choo},
  \textbf{JunSeung Lee},
  \textbf{Jimyeong Kim},
  \textbf{Yeeho Song},
  \textbf{S. K. Hong},
  \textbf{Yeong-Dae Kwon}
\\
  Samsung SDS
\\
  \small{\{jinho12.choo, juns2.lee, jimy.kim, yeeho.song, s.k.hong, y.d.kwon\}@samsung.com
  }
}

\begin{document}
\maketitle

\begin{abstract}

Large language models (LLMs) demonstrate strong multilingual capabilities, yet often fail to consistently generate responses in the intended language, exhibiting a phenomenon known as \emph{language confusion}.
Prior mitigation approaches based on sequence-level fine-tuning, such as DPO, ORPO, and GRPO, operate at the level of entire responses and can lead to unintended degradation of general model capabilities, motivating the need for more fine-grained alternatives.
To address this, we introduce \textbf{Token-Level Policy Optimization (TLPO)}, a fine-tuning framework designed to mitigate language confusion through localized, token-level updates. 
TLPO identifies error-prone positions, explores alternative candidate tokens, and updates the policy using a tailored objective to suppress error-inducing outputs at a granular level.
This selective intervention enables effective mitigation of language confusion without compromising the model's general abilities.
Experiments on multiple multilingual LLMs across diverse languages demonstrate that TLPO significantly outperforms baselines in improving language consistency while preserving downstream task accuracy.

\end{abstract}

\section{Introduction}

Large language models (LLMs) have demonstrated exceptional capabilities across diverse natural language processing tasks, driving their adoption in numerous applications~\cite{openai2023gpt4,zhao2023survey}. While a performance gap has historically existed between high-resource and low-resource languages~\cite{hu2020xtreme}, the emergence of open-weight multilingual LLMs—e.g., Llama~4, Qwen~3, and Aya—has substantially narrowed this gap, showing steady improvements in multilingual performance~\cite{grattafiori2024llama3,meta2025llama4,yang2025qwen3,team2025gemma,jiang2024mixtralexperts,bloom,aya_model}.

\begin{figure}[!t]    
    \centering
    \begin{subfigure}{0.45\textwidth} 
        \centering
        \includegraphics[width=\textwidth]{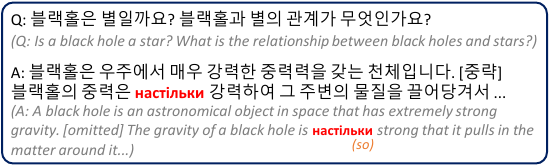}
        \caption{An example of language confusion. The Llama-3.1-8B-Instruct model generates a response to a Korean prompt that inadvertently includes Ukrainian words.}
        \label{fig:fig_lc_example}
    \end{subfigure}
    \vspace{7pt}
    
    \begin{subfigure}{0.48\textwidth}
        \centering
        \includegraphics[width=\textwidth]{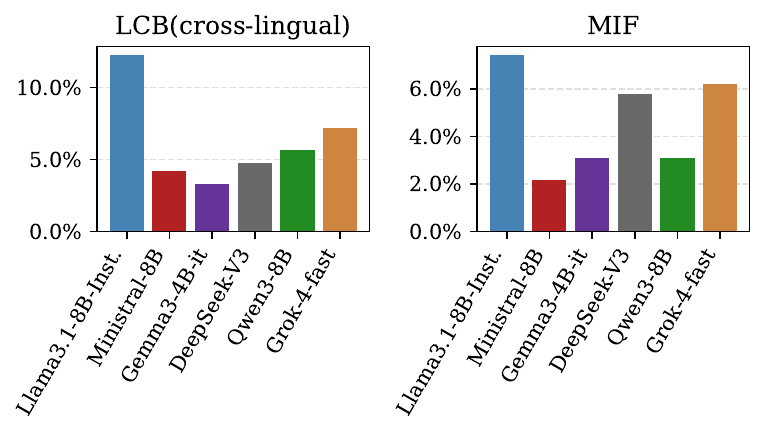}
        \caption{Proportion of responses exhibiting language confusion in LCB (cross-lingual)~\cite{marchisio2024understanding} and MIF~\cite{zeng2024marcobenchmif} tasks across various models.}
        \label{fig:fig_current_model_lc}
    \end{subfigure}
    \caption{Overview of language confusion in recent models.}
    \label{fig:fig_intro}
    \vspace{-7pt}
\end{figure}

Despite these advancements, \emph{language confusion}—where a model inadvertently mixes languages or shifts the target language entirely—remains a persistent issue in practical deployment~\cite{marchisio2024understanding,oh2025evaluating,nie2025mechanistic,lin2025languageconfusiongate,lee2025controlling}. Multilingual LLMs, which share parameters across languages, are particularly prone to this problem due to the curse of multilinguality, where capacity competition induces cross-lingual interference~\cite{conneau2020unsupervised}. Figure~\ref{fig:fig_intro} illustrates a representative example of such confusion and its prevalence across recently released models. This inconsistency undermines response reliability, posing a significant barrier to the effective deployment of real-world applications.

\citet{marchisio2024understanding} employed Supervised Fine-Tuning (SFT) to mitigate this issue. However, SFT typically necessitates extensive high-quality data and carries the risk of catastrophic forgetting, which can degrade general capabilities~\cite{kirkpatrick2017overcoming, luo2023empirical}. Alternatively, preference-based alignment methods such as DPO, GRPO, and ORPO optimize models using sequence-level rankings~\cite{rafailov2023dpo,shao2024deepseekmath,hong2024orpo, lee2025controlling}. Yet, these sequence-level objectives face inherent limitations; by treating the entire response as a monolithic unit, they lack the granularity to penalize specific error-inducing tokens without suppressing the valid surrounding context. This coarse-grained approach often necessitates a trade-off between rectifying localized errors and maintaining overall response quality, analogous to findings in mathematical reasoning where process-level supervision outperforms outcome-based metrics~\cite{lightman2023lets}.

In this paper, we introduce \textbf{Token-Level Policy Optimization (TLPO)}, a fine-tuning framework designed to precisely rectify localized errors, such as language confusion. Unlike sequence-level methods, TLPO identifies error-prone positions, explores alternative candidate tokens, and updates the policy using a tailored objective to suppress undesirable outputs at the token level. This granular strategy allows the model to eliminate errors effectively while preserving existing knowledge.

To the best of our knowledge, this is the first work to address language confusion by performing exploration and policy updates specifically at the positions where errors occur.

The key contributions of this paper are as follows:
 
\begin{itemize}

\item We introduce \emph{Token-Level Policy Optimization (TLPO)}, a fine-tuning framework designed to rectify localized errors. In contrast to coarse-grained sequence-level fine-tuning methods, TLPO enables precise policy updates by exploring and optimizing candidate tokens at error-prone positions.

\item We propose a \emph{probability-ranked exploration strategy} combined with a \emph{tailored advantage formulation}. This mechanism effectively suppresses error-inducing tokens locally, thereby addressing specific issues without degrading the model's general capabilities.

\item We demonstrate the efficacy of TLPO in mitigating \emph{language confusion}. Through extensive experiments on diverse multilingual LLMs, we show that our approach significantly outperforms sequence-level baselines in reducing confusion rates while maintaining performance on downstream tasks.

\end{itemize}

\section{Related Work}

\subsection{Preference-Based Fine-tuning of Large Language Models}

A foundational approach for aligning Large Language Models (LLMs) with human intent is \emph{Reinforcement Learning from Human Feedback (RLHF)}. \citet{christiano2017deep} introduced the paradigm of training a reward model from pairwise preferences and optimizing the policy via reinforcement learning. Building on this, \citet{ouyang2022training} aligned GPT-3 into InstructGPT through a three-stage pipeline comprising supervised fine-tuning (SFT), reward model training, and PPO-based optimization.

To mitigate the complexity and instability inherent in PPO-style RLHF, reward-free preference optimization methods have emerged. \citet{rafailov2023dpo} proposed \emph{Direct Preference Optimization (DPO)}, which derives a closed-form solution to the KL-regularized objective, enabling preference learning via a pairwise logistic loss without an explicit reward model. Subsequent methods, such as ORPO~\cite{hong2024orpo} and KTO~\cite{ethayarajh2024kto}, further integrate preference signals directly into the SFT objective, thereby reducing reliance on reference models or paired data.

Concurrently, RL-based methods continue to evolve toward greater efficiency. DeepSeekMath~\cite{shao2024deepseekmath} and DeepSeek-R1~\cite{guo2025deepseek} introduced \emph{Group Relative Policy Optimization (GRPO)}, a critic-free approach that computes advantages from relative rewards among multiple outputs generated from a single prompt. This method reduces computational overhead and avoids value-function approximation biases, demonstrating effectiveness in mathematical reasoning.

Recently, approaches optimizing alignment signals at the token level have emerged to address the limitations of sequence-level preference optimization, specifically the challenge of precise credit assignment. 
For instance, \citet{xu2024finegrained} proposed a method that identifies preference-determining tokens by minimally editing rejected responses. Based on this data, they train a token-level reward model to update the policy via fine-grained PPO. 
Meanwhile, \citet{zhang2025tokenlevelacceptrejectmicro} propose a micro-alignment framework that bypasses parameter updates for the main model. Instead, it employs a lightweight external module that operates independently of the base LLM and intervenes during decoding, dynamically “accepting” or “rejecting” candidate tokens to facilitate alignment.

\subsection{Analysis and Mitigation of Language Confusion}

\citet{marchisio2024understanding} proposed the Language Confusion Benchmark (LCB) to quantify confusion severity across diverse conditions. Extending this, \citet{oh2025evaluating} introduced evaluation settings that include code-switching scenarios, capturing language-selection failures in realistic conversational contexts.

In terms of mitigation, \citet{marchisio2024understanding} demonstrated that multilingual instruction tuning and inference-time controls (e.g., few-shot prompting) can reduce confusion. More recently, \citet{lee2025controlling} applied ORPO-based fine-tuning using pairs of target-language responses (chosen) and code-mixed responses (rejected).
However, as their analysis focused solely on QA benchmarks, the impact on general capabilities across a wider range of tasks has not been fully investigated.
Furthermore, sequence-level optimization may inadvertently suppress valid information contained within a \emph{rejected} response, leading to misguided policy updates.

Mechanistic interpretability offers an alternative perspective by identifying internal causes of confusion. \citet{nie2025mechanistic} analyzed \emph{confusion points} and associated components (e.g., attention heads) to propose neuron editing. Similarly, \citet{lin2025languageconfusiongate} introduced a Language Confusion Gate to suppress inconsistent tokens during decoding. While promising, these methods often require model-specific heuristics or invasive modifications to internal mechanisms, limiting their scalability and ease of deployment compared to training-based approaches.

\section{Methods}

\begin{figure}[t] 
    \centering
    
    \begin{subfigure}[b]{0.47\textwidth}
        \centering
        \includegraphics[width=\linewidth]{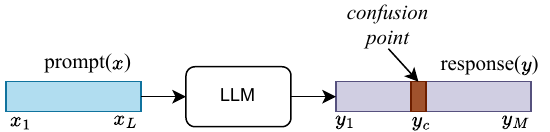} 
        \caption{Detect confusion point $c$.}
        \label{fig:sub_a}
    \end{subfigure}
    
    \vspace{8pt}
    
    \begin{subfigure}[b]{0.47\textwidth}
        \centering
        \includegraphics[width=\linewidth]{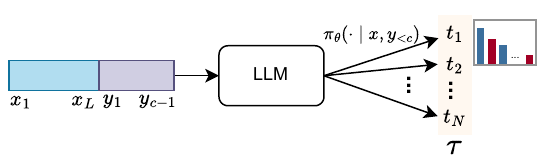}
        \caption{Obtain candidate tokens $\mathcal{T}$ at the confusion point $c$.}
        \label{fig:sub_b}
    \end{subfigure}
    
    \vspace{8pt}
    
    \begin{subfigure}[b]{0.47\textwidth}
        \centering
        \includegraphics[width=\linewidth]{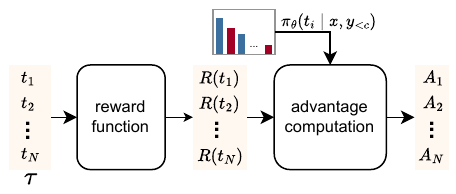}
        \caption{Compute advantage $A_i$ for each candidate token.}
        \label{fig:sub_c}
    \end{subfigure}

    \caption{Overview of the Token-Level Policy Optimization (TLPO) framework.}
    \label{fig:main}
    \vspace{-5pt}
\end{figure}

\textbf{Token-Level Policy Optimization (TLPO)} operates by precisely identifying token positions requiring adjustment, exploring alternative candidates at these locations, and subsequently optimizing the policy. 
Its primary goal is to suppress error-inducing tokens such as language confusion without compromising the LLM's inherent knowledge. 
By pinpointing the exact confusion points and deriving training signals exclusively from these instances—rather than optimizing the entire response sequence—TLPO minimizes the risk of global performance degradation. Furthermore, we employ a tailored objective function to ensure that policy updates are strictly targeted; this enables the effective elimination of \emph{language confusion} while preserving the LLM's original generative capabilities.
Figure~\ref{fig:main} illustrates the overall framework of TLPO.

\subsection{Probability-Ranked Token Exploration}
\label{subsec:tlpo_exploration}
We define a large language model (LLM) $\pi_{\theta}$, parameterized by $\theta$, as a conditional probability distribution $\pi_{\theta}(y_t \mid x, y_{<t})$ that predicts the next token $y_t$ given a prompt $x$ and preceding tokens $y_{<t}$. Let $y = [y_1, y_2, \dots, y_T]$ denote an output sequence where each token $y_t$ belongs to a vocabulary $\mathcal{V}$. The conditional probability of the entire sequence $y$ is given by $\pi_{\theta}(y \mid x) = \prod_{t=1}^T \pi_{\theta}(y_t \mid x, y_{<t})$.

Given a prompt $x$ sampled from the training dataset $\mathcal{D}$, we first generate a response sequence $y$ by autoregressively executing $\pi_{\theta}$. Responses entirely free of \emph{language confusion} provide no error signal and are consequently excluded from the training phase. Conversely, if $y$ exhibits \emph{language confusion}, we identify the \emph{confusion point} $c$, defined as the index of the first token decoded in a language other than the target language, following \citet{nie2025mechanistic}. Further details on the confusion detection process are provided in Appendix~\ref{sec:appendix_detect_lc_detail}.

At the identified confusion point $c$, we employ a \textbf{probability-ranked exploration} strategy that prioritizes the most probable next-token candidates. Given the distribution $\pi_{\theta}(\cdot \mid x, y_{<c})$, we select the top-$N$ ($N \ge 2$) tokens with the highest probabilities to form the candidate set $\mathcal{T}$:
\begin{equation}
\label{eq:def_T}
\begin{aligned}
\mathcal{T}(x, y_{<c}) &= \{ t_i \mid i \in \mathcal{I}_N(x, y_{<c}) \}, \\
\mathcal{I}_N(x, y_{<c})
&= \operatorname*{arg\,topN}_{i \in \{1,\dots,|\mathcal{V}|\}}
\pi_{\theta}(t_i \mid x, y_{<c}),
\end{aligned}
\end{equation}
where $\operatorname*{arg\,topN}$ returns the indices of the $N$ tokens with the largest probabilities, sorted in descending order.

By focusing evaluation and optimization on these high-probability candidates, TLPO concentrates parameter updates on tokens most likely to be generated at the confusion point. This enables efficient suppression of erroneous outputs. Moreover, as discussed in Section~\ref{subsec:after_topn}, we observe that updating parameters using only $\mathcal{T}$ implicitly reduces the probabilities of confusion-inducing tokens outside the set $\mathcal{T}$. We conjecture that this phenomenon arises from the presence of language-specific components within the model~\cite{nie2025mechanistic}; suppressing a subset of tokens associated with a particular language concurrently dampens the activation of other tokens belonging to the same language.

\subsection{Optimization Objective}
\label{subsec:tlpo}
In the policy optimization phase, we update the policy parameters $\theta$ using the candidate set $\mathcal{T}$
to suppress error-inducing tokens without compromising pre-existing knowledge. 
This section details the objective function and the specialized advantage formulation designed for this purpose.

Fine-tuning the policy $\pi_\theta$ is formulated as maximizing the expected reward ${J}(\theta)$.
Based on the reward $R(y)$, the sequence-level objective ${J}(\theta)$ is defined as:
\begin{equation}
\label{eq:def_J}
  J(\theta)
  = \mathbb{E}_{x \sim \mathcal{D},\, y \sim \pi_{\theta}(\cdot \mid x)}
  \bigl[ R(y) \bigr] .
\end{equation}

Here, $R(y)$ is a function that yields a reward value based on whether \textit{language confusion} occurs in the LLM's response sequence $y$.

TLPO approximates this sequence-level improvement by maximizing the expected reward of the candidate tokens $\mathcal{T}$ specifically at the confusion point. The resulting token-level objective for TLPO is formulated as follows:
\begin{equation}
\label{eq:tlpo_j_pre}
  {J}_{\mathrm{TLPO}}(\theta) 
  = \mathbb{E}_{x \sim \mathcal{D},\, y \sim \pi_{\theta}(\cdot \mid x)}
    \Biggl[
      \frac{1}{N} \sum_{t_i \in \mathcal{T}} R(t_i)
    \Biggr].
\end{equation}

To optimize this, we adapt the PPO objective~\cite{schulman2017proximal} to our setting:
{
\fontsize{10.4pt}{11pt}\selectfont
\begin{equation}
\label{eq:tlpo_j}
\begin{aligned}
  &{J}_{\mathrm{TLPO}}(\theta)  \
    =   \mathbb{E}_{x \sim \mathcal{D}, y \sim \pi_{\theta_{\mathrm{old}}}(\cdot \mid x)} \\
  &\quad\quad\quad \Biggl[
    \frac{1}{N}
    \sum_{t_i \in \mathcal{T}}
    \Bigl(\min\Bigl(
      \frac 
        {\pi_{\theta}(t_i \mid x, y_{<c})} 
        {\pi_{\theta_{\mathrm{old}}} (t_i \mid x, y_{<c})}
      {A}_{i},\, \\      
      &\quad\quad\quad\quad \operatorname{clip}\bigl(
        \frac 
            {\pi_{\theta}(t_i \mid x, y_{<c})} 
            {\pi_{\theta_{\mathrm{old}}}(t_i \mid x, y_{<c})},
        1-\varepsilon,\,
        1+\varepsilon
      \bigr)\,
      {A}_{i}
    \Bigr)
   \\
  &\quad\quad\quad\quad  -\, 
  \beta\,
  D_{\mathrm{KL}}\!\bigl(
    \pi_{\theta}
    \,\Vert\,
    \pi_{\theta_{\mathrm{ref}}}
  \bigr) \Bigr) \Biggr].
\end{aligned}
\end{equation}
}

Here, $\pi_{\theta_{\mathrm{old}}}$ refers to the policy under which the candidate set $\mathcal{T}$ was selected, which is the model policy before the current update step, whereas $\pi_{\theta_{\mathrm{ref}}}$ represents the initial policy before applying TLPO.

We design the advantage function to reflect our probability-ranked exploration strategy: 
\begin{equation}
\label{eq:def_A}
  A_i = \frac{1}{Z} \cdot
  {
    \pi_{\theta_{\mathrm{old}}}(t_i \mid x, y_{<c}) \bigl( R(t_i) - \mu \bigr)
  },
\end{equation}
where 
\begin{equation}
\label{eq:def_mu}
\begin{aligned}
  \mu & = 
    \frac{\sum_{j=1}^N \bigl( \pi_{\theta_{\mathrm{old}}}(t_j \mid x, y_{<c}) R(t_j) \bigr)}
       {\sum_{j=1}^N \pi_{\theta_{\mathrm{old}}}(t_j \mid x, y_{<c})}, \\
  Z & = \sum_{j=1}^N \bigl| \pi_{\theta_{\mathrm{old}}}(t_j \mid x, y_{<c}) 
    \bigl( R(t_j) - \mu \bigr)
    \bigr|.
\end{aligned}
\end{equation}

Since our exploration process deterministically selects candidates $\mathcal{T}$ based on probability rank rather than through sampling, we incorporate the original token probability into the advantage formulation by multiplying it with the centered reward term $(R(t_i) - \mu)$. This formulation ensures that the advantage scales in proportion to the original probabilities within both the positive and negative reward token sets, respectively. Such a design encourages the model to maintain the relative probability distribution of valid tokens even after the suppression of error-inducing ones, thereby preserving the LLM's originally learned distribution as much as possible.

\begin{figure}[tb]
\centering
\includegraphics[width=0.97\columnwidth]{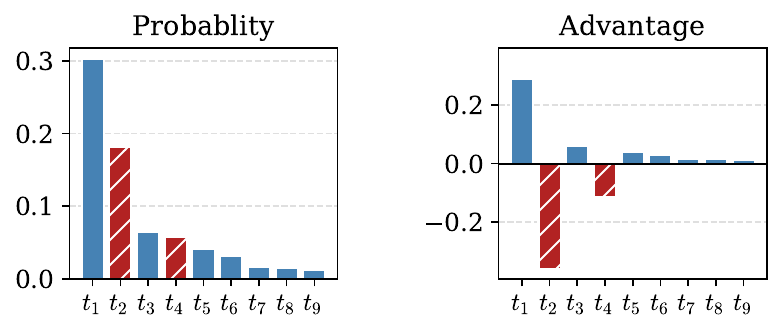}
\caption{An example of the advantage distribution. The red hatched bars represent the probabilities and advantages of confusion-inducing tokens.}
\label{fig:fig_prob_adv}
\vspace{-8pt}
\end{figure}

Here, $\mu$ represents the probability-weighted average of the token rewards. And $Z$ serves as a normalization constant. By ensuring that the sum of the absolute values of the advantages across all candidate tokens equals $1$, $Z$ maintains a consistent scale for the training signals regardless of variations in raw probabilities or rewards, thereby enhancing the stability of the optimization process.
Figure~\ref{fig:fig_prob_adv} illustrates the relationship between token probability and the calculated advantage.

For the KL divergence term $D_{\mathrm{KL}}$ in Equation~\eqref{eq:tlpo_j}, we employ the unbiased estimator proposed in~\cite{schulman2020kld}, as in GRPO:
\begin{equation}
\label{eq:def_lkd}
\begin{aligned}
  D_{\mathrm{KL}}\!\bigl(
    \pi_{\theta}
    \,\Vert\,
    \pi_{\theta_{\mathrm{ref}}}
   \bigr) 
   & =
    \frac 
        {\pi_{\theta_{\mathrm{ref}}}(t_i \mid x, y_{<c})}
        {\pi_{\theta}(t_i \mid x, y_{<c})} \\
    & - \log
    \frac
        {\pi_{\theta_{\mathrm{ref}}}(t_i \mid x, y_{<c})}
        {\pi_{\theta}(t_i \mid x, y_{<c})}
    - 1.
\end{aligned}    
\end{equation}

The reward function $R(t_i)$ yields a reward value for the token $t_i$ based on whether it leads to \emph{language confusion}.
To accurately assess whether $t_i$ contributes to language confusion upon detokenization, we generate a short lookahead sequence of $k$ additional tokens.
$k$ should be one less than the maximum number of tokens required to represent a single character. In practice, we set a small lookahead of $k=3$, which we found sufficient for the tokenizers used in our experiments.
These $k$ additional tokens are generated autoregressively following the distribution $\pi_{\theta}( \cdot \mid x, y_{<c}, t_i)$. 
Finally, we decode the concatenation of $t_i$ and the lookahead tokens to verify the occurrence of language confusion, and determine the reward for $t_i$ accordingly.

In summary, TLPO enables targeted fine-tuning by identifying correction points, evaluating multiple candidates, and optimizing parameters to effectively eliminate \emph{language confusion} while minimizing general performance degradation. The complete algorithm is presented in Appendix~\ref{sec:appendix_tlpo_algorithm}.

\section{Experiments}

\subsection{Experimental Setup}

\paragraph{Target Languages and Base Models}
We evaluate the effectiveness of TLPO in mitigating \textit{language confusion} across four target languages: Chinese, Arabic, Korean, and Japanese. The base models employed in our experiments are Llama-3.1-8B-Instruct, Qwen3-8B, Ministral-8B-Instruct, and Gemma-3-4B-IT. 

For fine-tuning, we utilize the training split of Bactrian-X, a multilingual instruction-following dataset~\cite{li2023bactrianx}. Detailed specifications regarding the training data composition are provided in Appendix~\ref{sec:appendix_train_dataset}.

\paragraph{Baselines and Evaluation Benchmarks}

For comparative analysis, we employ Supervised Fine-Tuning (SFT) and sequence-level preference optimization methods, specifically DPO~\cite{rafailov2023dpo} and ORPO~\cite{hong2024orpo}, as baseline methods\footnote{The experiments for the baseline methods were conducted using the TRL library. 
We conducted GRPO experiments by assigning a reward of +1 for responses free of language confusion and -1 for those exhibiting confusion. 
However, we observed a progressive reduction in response length as fine-tuning advanced. Due to this instability, GRPO results were excluded from our analysis.}.

Our evaluation is twofold: \textit{language confusion} assessment and general accuracy assessment.
We evaluate \textit{language confusion} on MIF, MMMLU, LCB-crosslingual, LCB-monolingual, and GSM8K(cross\footnote{Problems are presented in English, while the instructions require generating the solution in the target language. Please refer to Appendix~\ref{sec:appendix_eval_dataset} for detailed specifications.})~\cite{zeng2024marcobenchmif,openai2024mmmlu,marchisio2024understanding,cobbe2021gsm8k}.
To assess general task performance, we employ MIF(English/target\footnote{Here, `target' denotes the target language dataset.}), MMLU, MMMLU(target), GPQA, GPQA-diamond, ARC-Challenge, Big-Bench-Hard, MATH, and GSM8K(English/cross)~\cite{hendryckstest2021,rein2023gpqa,clark2018arc,suzgun2022challenging,hendrycks2021math,cobbe2021gsm8k}. 
Further details on evaluation settings are described in Appendix~\ref{sec:appendix_eval_dataset}.

\paragraph{Evaluation Scenarios for English Tokens}
In this study, we conduct our experiments under two distinct settings regarding the treatment of English tokens: one where English is classified as a \textit{neutral category}, and another where any non-target English generation is strictly treated as \textit{language confusion}.

In the first setting, English is treated as belonging neither to the target language nor to the confused language, and its presence is not penalized. This neutral treatment is motivated by the fact that English is naturally intermixed in diverse linguistic environments—appearing in abbreviations, domain-specific terminology, and structural markers such as section headers. Thus, this approach aligns more closely with real-world deployment scenarios~\cite{marchisio2024understanding,nie2025mechanistic}. Crucially, the generation of English often serves to maintain semantic precision; consequently, indiscriminately classifying English instances as confusion can distort the model's knowledge representation, potentially leading to a degradation in accuracy.

Nonetheless, for a more rigorous validation, we include a stricter scenario that treats any English output as confusion. Evaluating the proposed methodology and baselines across these two criteria—encompassing both real-world usage and strict language adherence—ensures a comprehensive and reliable measure of their alignment performance.

\paragraph{Implementation Details}
All experiments were conducted on a single node equipped with eight NVIDIA H100 GPUs.

The source code for our experiments is available at \url{https://github.com/samsungsds-research-papers/TLPO}.

\begin{table*}[!ht]
    \centering
    
    \begin{subtable}{0.98\textwidth}
    \centering
    \fontsize{8.5pt}{9pt}\selectfont
    \begin{tabular}{c||ccccc|c}
    \toprule
    \multirow{2}{*}{\textbf{Method}} & LCB & LCB & MIF & MMMLU & GSM8K  & \multirow{2}{*}{\textbf{Mean}} \\
    & (cross-lingual) & (monolingual) & (target) & (target) & (cross)  & \\
    \midrule 
    \midrule 
    \textbf{Baseline} & 93.66(99.87) & 96.12(99.97) & 97.52(99.97) & 98.15(99.89) & 97.97(99.92) & 96.68(99.92) \\
    \midrule 
    \textbf{SFT} & 97.41(99.90) & \textbf{99.90(100.00)} & \textbf{99.54(99.96)} & 99.11(99.75) & \textbf{99.72(99.99)} & 99.14(99.92) \\
    \textbf{DPO} & 96.31(99.65) & 98.37(99.87) & 98.92(99.83) & 98.94(99.43) & 99.02(99.83) & 98.31(99.72) \\
    \textbf{ORPO} & 94.35(99.80) & 97.51(99.97) & 98.03(99.91) & 97.85(99.82) & 98.63(99.90) & 97.27(99.88) \\
    \textbf{TLPO(ours)} & \textbf{97.68(99.97)} & 99.72(100.00) & 99.46(99.99) & \textbf{99.49(99.96)} & 99.58(99.99) & \textbf{99.19(99.98)} \\    
    \bottomrule         
    \end{tabular}

    \caption{Average Response Pass Rate (RPR) and Word Pass Rate (WPR). Values are presented as RPR(WPR). All values are in percentages.}        
    \label{tab:main_result_a}
    \end{subtable}

    \par\bigskip

    \begin{subtable}{0.98\textwidth}
    \centering
    \fontsize{8.2pt}{9pt}\selectfont
    \begin{tabular}{c||cccccccccc|c}    
    \toprule
         \multirow{2}{*}{\fontsize{8pt}{9pt}\selectfont \textbf{Method}} 
         & \fontsize{8pt}{9pt}\selectfont MIF
         & \fontsize{8pt}{9pt}\selectfont MIF
         & \fontsize{7.2pt}{9pt}\selectfont MMMLU
         & \fontsize{8pt}{9pt}\selectfont GPQA
         & \fontsize{8pt}{9pt}\selectfont GPQA
         & \fontsize{7.2pt}{9pt}\selectfont ARC-C
         & \fontsize{8pt}{9pt}\selectfont BBH
         & \fontsize{8pt}{9pt}\selectfont MATH
         & \fontsize{7.2pt}{9pt}\selectfont GSM8K
         & \fontsize{7.2pt}{9pt}\selectfont GSM8K
         & \multirow{2}{*}{\fontsize{8pt}{9pt}\selectfont \textbf{Mean}} \\
         
         & \fontsize{8pt}{9pt}\selectfont (en)       
         & \fontsize{8pt}{9pt}\selectfont (target) 
         & \fontsize{8pt}{9pt}\selectfont (target) 
         & \fontsize{8pt}{9pt}\selectfont (en)  
         & \fontsize{7.1pt}{9pt}\selectfont (diamond, en)
         & \fontsize{8pt}{9pt}\selectfont (en)  
         & \fontsize{8pt}{9pt}\selectfont (en)      
         & \fontsize{8pt}{9pt}\selectfont (en)  
         & \fontsize{8pt}{9pt}\selectfont (en)  
         & \fontsize{8pt}{9pt}\selectfont (cross) 
         &  \\
    \midrule 
    \midrule 
        \textbf{Baseline} & 69.66 & 50.47 & 55.07 & 33.66 & 32.54 & 82.55 & 50.07 & 49.46 & 78.48 & 81.52 & 58.35 \\
    \midrule 
        \textbf{SFT} & 61.14 & 39.91 & 46.54 & 28.24 & 27.87 & 82.81 & \textbf{56.52} & 41.35 & 59.35 & 63.34 & 50.71 \\
        \textbf{DPO} & 67.31 & 46.68 & 52.76 & 30.85 & 31.53 & 82.59 & 49.21 & 44.79 & 75.06 & 78.57 & 55.94 \\
        \textbf{ORPO} & 66.00 & 43.82 & 51.58 & 30.52 & 31.06 & \textbf{82.99} & 49.22 & 44.73 & 73.14 & 78.09 & 55.12 \\
        \textbf{TLPO(ours)} & \textbf{69.22} & \textbf{49.71} & \textbf{54.24} & \textbf{31.58} & \textbf{33.87} & 82.52 & 50.90 & \textbf{48.21} & \textbf{79.55} & \textbf{80.99} & \textbf{58.08} \\
    \bottomrule         
    \end{tabular}

    \caption{Average accuracy after fine-tuning.}    
    \label{tab:main_result_b}
    \end{subtable}

    \caption{
        Performance comparison of TLPO against baselines (SFT~\cite{marchisio2024understanding}, DPO~\cite{rafailov2023dpo}, and ORPO~\cite{lee2025controlling}) under English as a Neutral Category. Results are reported as average RPR, WPR, and accuracy across four models and four target languages. For TLPO, we set $N=16$. Detailed results are provided in Appendix~\ref{sec:appendix_detail_result}.        
    }
    
    \label{tab:main_result}
    \vspace{-3pt}    
\end{table*}

\begin{figure*}[!tb]    
    \centering
    \includegraphics[width=0.98\linewidth]{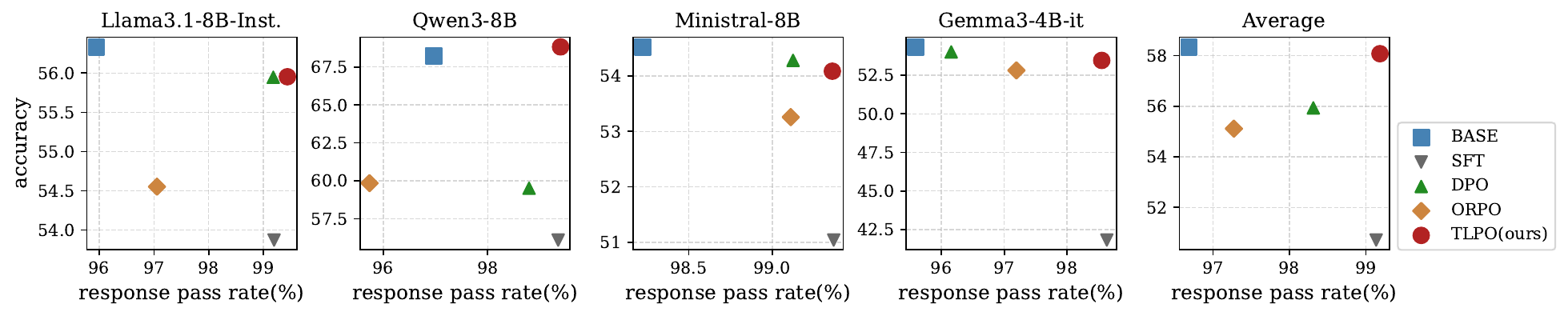}

    \caption{
        Scatter plot of the average Response Pass Rate (RPR) versus accuracy for each method after fine-tuning under English as a Neutral Category. BASE denotes the original model prior to fine-tuning. Detailed results are provided in Appendix~\ref{sec:appendix_detail_result}.
    }
    \label{fig:result_main}
    \vspace{-5pt}
\end{figure*}

\subsection{Definition of Language Confusion Metrics}

To quantitatively evaluate \emph{language confusion}, we employ two metrics: \emph{Word Pass Rate} (WPR) and \emph{Response Pass Rate} (RPR).

\paragraph{Word Pass Rate (WPR)}

WPR denotes the proportion of \emph{non-confused words} relative to the total number of words generated by the LLM. Here, a "non-confused word" is defined as a word in which all constituent characters belong to the character set of the target language.

Equation~\eqref{eq:def_WPR} presents the formulation of WPR, which aligns with the definitions used in prior studies~\cite{marchisio2024understanding}.
\begin{equation}
\label{eq:def_WPR}
    \mathrm{WPR} = \frac{|\mathcal{W}_{pass}|}{|\mathcal{W}_{total}|},
\end{equation}
where
$\mathcal{W}_{total}$ denotes the set of all generated words, 
and $\mathcal{W}_{pass} = \{w \in \mathcal{W}_{total} \mid w \text{ is non-confused} \}$.

\paragraph{Response Pass Rate (RPR)}

RPR indicates the proportion of \emph{non-confused responses} out of the total responses generated by the LLM for a given evaluation dataset. A "non-confused response" is defined as a response sequence that is entirely free of words exhibiting language confusion.

RPR is defined as follows:
\begin{equation}
\label{eq:def_RPR}
    \mathrm{RPR} = \frac{|\mathcal{R}_{pass}|}{|\mathcal{R}_{total}|},
\end{equation}
where
$\mathcal{R}_{total}$ denotes the set of all generated responses, and 
$\mathcal{R}_{pass} = \{r \in \mathcal{R}_{total} \mid r \text{ is non-confused} \}$.


\subsection{Experimental Results on Mitigating Language Confusion}
\label{sec:main_result}

In this section, we analyze the performance of TLPO and baselines under the two evaluation scenarios based on the treatment of English tokens: the neutral category setting and the strict confusion setting.


\subsubsection{Results under English as a Neutral Category}

Table~\ref{tab:main_result} summarizes the quantitative results obtained under the neutral English treatment. As shown in Table~\ref{tab:main_result}(\subref{tab:main_result_a}), all fine-tuning methods improve the Response Pass Rate (RPR) compared to the baseline (96.68\%). Notably, TLPO achieves the highest average RPR of 99.19\%, effectively mitigating language confusion across all evaluated benchmarks. While SFT also demonstrates strong mitigation capabilities with an average of 99.14\%, preference-based methods like DPO and ORPO show relatively lower effectiveness.

Table~\ref{tab:main_result}(\subref{tab:main_result_b}) presents the accuracy across various downstream tasks after fine-tuning for each method, thereby quantifying the extent of general performance degradation caused by language confusion mitigation.
SFT suffers from severe performance degradation, with the mean accuracy dropping from 58.35\% (Baseline) to 50.71\%, indicating a significant loss of general knowledge during the alignment process. DPO and ORPO also exhibit notable declines, resulting in accuracies of 55.94\% and 55.12\%, respectively.
In contrast, TLPO successfully preserves the model's general capabilities, achieving a mean accuracy of 58.08\%. This performance is comparable to the baseline and consistently outperforms other fine-tuning methods across most benchmarks, demonstrating that TLPO mitigates language confusion without compromising the model's reasoning and knowledge retrieval abilities.

Figure~\ref{fig:result_main} displays the relationship between average RPR and average accuracy after fine-tuning under the neutral category setting.
This plot provides an immediate view of the shifts in RPR and accuracy induced by each method.
Here, we observe that TLPO consistently improves RPR while effectively minimizing accuracy degradation across all models.

In summary, extensive experiments across diverse models and tasks demonstrate that TLPO provides the most effective mitigation of language confusion while minimizing the loss of general knowledge. While SFT leads to the most substantial decline in accuracy across downstream tasks, and preference-based methods such as DPO and ORPO yield suboptimal compromises, TLPO’s token-level optimization precisely resolves linguistic issues without compromising the model’s general abilities. This establishes TLPO as a highly effective methodology for multilingual alignment, capable of selectively correcting errors without eroding core model competencies.

\begin{table*}[!htbp]
    \centering
    
    \begin{subtable}{0.98\textwidth}
    \centering
    \fontsize{8.5pt}{9pt}\selectfont
    \begin{tabular}{c||ccccc|c}
    \toprule
    \multirow{2}{*}{\textbf{Method}} & LCB & LCB & MIF & MMMLU & GSM8K  & \multirow{2}{*}{\textbf{Mean}} \\
    & (cross-lingual) & (monolingual) & (target) & (target) & (cross)  & \\
    \midrule 
    \midrule 
    \textbf{Baseline} & 43.88(78.05) & 83.48(99.65) & 84.94(98.41) & 43.66(58.26) & 60.42(77.17) & 63.27(82.31) \\
    \midrule 
    \textbf{SFT} & 38.33(64.37) & 62.75(98.12) & 60.26(95.03) & 44.61(55.20) & 30.05(52.33) & 47.20(73.01) \\
    \textbf{DPO} & 58.28(81.59) & 87.91(95.92) & 86.68(95.19) & 55.64(65.44) & 75.13(81.96) & 72.73(84.02) \\
    \textbf{ORPO} & 50.37(82.34) & 83.72(96.26) & 84.41(95.44) & \textbf{57.86(71.43)} & 72.37(87.07) & 69.75(86.51) \\
    \textbf{TLPO(ours)} & \textbf{57.25(78.66)} & \textbf{96.44(99.90)} & \textbf{96.66(99.73)} & 52.25(60.33) & \textbf{85.34(89.58)} & \textbf{77.59(85.64)} \\
        \bottomrule         
    \end{tabular}

    \caption{Average Response Pass Rate (RPR) and Word Pass Rate (WPR). Values are presented as RPR(WPR). All values are in percentages.}        
    \label{tab:main_result_ief_a}
    \end{subtable}

    \par\bigskip

    \begin{subtable}{0.98\textwidth}
    \centering
    \fontsize{8.2pt}{9pt}\selectfont
    \begin{tabular}{c||cccccccccc|c}    
    \toprule
         \multirow{2}{*}{\fontsize{8pt}{9pt}\selectfont \textbf{Method}} 
         & \fontsize{8pt}{9pt}\selectfont MIF
         & \fontsize{8pt}{9pt}\selectfont MIF
         & \fontsize{7.2pt}{9pt}\selectfont MMMLU
         & \fontsize{8pt}{9pt}\selectfont GPQA
         & \fontsize{8pt}{9pt}\selectfont GPQA
         & \fontsize{7.2pt}{9pt}\selectfont ARC-C
         & \fontsize{8pt}{9pt}\selectfont BBH
         & \fontsize{8pt}{9pt}\selectfont MATH
         & \fontsize{7.2pt}{9pt}\selectfont GSM8K
         & \fontsize{7.2pt}{9pt}\selectfont GSM8K
         & \multirow{2}{*}{\fontsize{8pt}{9pt}\selectfont \textbf{Mean}} \\
         
         & \fontsize{8pt}{9pt}\selectfont (en)       
         & \fontsize{8pt}{9pt}\selectfont (target) 
         & \fontsize{8pt}{9pt}\selectfont (target) 
         & \fontsize{8pt}{9pt}\selectfont (en)  
         & \fontsize{7.1pt}{9pt}\selectfont (diamond, en)
         & \fontsize{8pt}{9pt}\selectfont (en)  
         & \fontsize{8pt}{9pt}\selectfont (en)      
         & \fontsize{8pt}{9pt}\selectfont (en)  
         & \fontsize{8pt}{9pt}\selectfont (en)  
         & \fontsize{8pt}{9pt}\selectfont (cross) 
         &  \\
    \midrule 
    \midrule 
    \textbf{Baseline} & 69.63 & 50.37 & 55.14 & 32.83 & 32.32 & 82.55 & 50.06 & 49.59 & 78.42 & 81.50 & 58.24 \\
    \midrule 
    \textbf{SFT} & 61.14 & 39.91 & 46.54 & 28.24 & 27.87 & \textbf{82.81} & \textbf{56.52} & 41.35 & 59.35 & 63.34 & 50.71 \\
    \textbf{DPO} & \textbf{69.26} & 43.77 & 48.37 & 31.10 & 31.09 & 82.32 & 48.66 & 42.00 & 75.04 & 74.44 & 54.60 \\
    \textbf{ORPO} & 65.72 & 42.62 & 50.18 & \textbf{31.52} & 31.19 & 82.77 & 48.93 & 43.78 & 72.86 & 76.50 & 54.61 \\
    \textbf{TLPO(ours)} & 65.76 & \textbf{46.21} & \textbf{53.83} & 30.66 & \textbf{31.66} & 82.29 & 46.97 & \textbf{47.73} & \textbf{77.86} & \textbf{78.71} & \textbf{56.17} \\
        \bottomrule         
    \end{tabular}

    \caption{Average accuracy after fine-tuning.}     
    \label{tab:main_result_ief_b}
    \end{subtable}

    \caption{        
        Performance comparison of TLPO ($N=16$) against SFT~\cite{marchisio2024understanding}, DPO~\cite{rafailov2023dpo}, and ORPO~\cite{lee2025controlling}, treating English occurrence as language confusion. Results are averaged (RPR, WPR, and accuracy) across four models and four target languages.
    }
    
    \label{tab:main_result_ief}
    \vspace{-3pt}    
\end{table*}

\begin{figure*}[!tb]    
    \centering
    \includegraphics[width=0.98\linewidth]{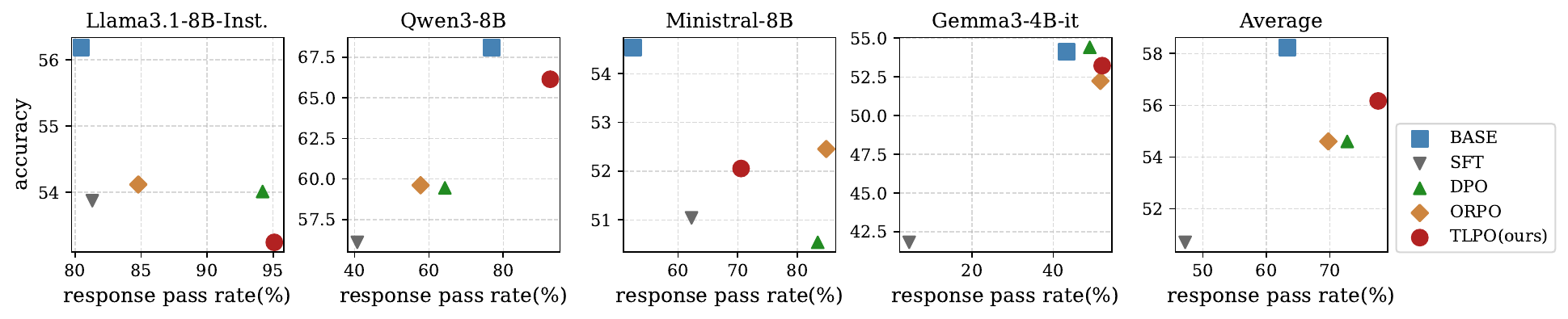}

    \caption{Scatter plot of the average Response Pass Rate (RPR) versus accuracy for each method after fine-tuning, treating English occurrence as language confusion. BASE denotes the original model prior to fine-tuning.
    }
    \label{fig:result_main_ief_all}
    \vspace{-5pt}
\end{figure*}

\subsubsection{Results under English as Language Confusion}

Table~\ref{tab:main_result_ief} and Figure~\ref{fig:result_main_ief_all} present the evaluation results under the stricter scenario where any English output is treated as language confusion. In this challenging setting, the Baseline RPR drops significantly to 63.27\%, reflecting the frequent occurrence of English tokens in standard LLM responses.

As shown in Table~\ref{tab:main_result_ief}(\subref{tab:main_result_ief_a}), SFT fails to improve language adherence, with its RPR further declining to 47.20\%. This suggests that enforcing strict language constraints through traditional SFT can lead to unstable alignment. While preference-based methods such as DPO (72.73\%) and ORPO (69.75\%) show improvements over the baseline, TLPO achieves the highest average RPR of 77.59\%. This confirms TLPO's robustness even under stringent linguistic constraints.

Regarding the general task performance presented in Table~\ref{tab:main_result_ief}(\subref{tab:main_result_ief_b}), all fine-tuning methods exhibit a notable decline in accuracy compared to the baseline (58.24\%). We attribute this to the distortion of the model's inherent knowledge representation when English—a primary language for reasoning and knowledge—is strictly suppressed. However, even in this environment, TLPO maintains the highest mean accuracy of 56.17\%, outperforming SFT (50.71\%), DPO (54.60\%), and ORPO (54.61\%). 

In conclusion, these results demonstrate that while strict English suppression inevitably harms model performance, TLPO provides the most favorable balance by achieving superior alignment precision with the least degradation in core model capabilities.

\subsection{Probability Shifts in Tokens Beyond the Top-$N$ Candidates}
\label{subsec:after_topn}

In this section, we analyze how the probabilities of tokens outside the top-$N$ set change when parameter updates are performed using only the selected top-$N$ tokens. To investigate this, we conducted a controlled experiment with $N = 8$, using 100 curated prompts in which exactly three of the top-8 tokens were confusion-inducing. Accordingly, parameter updates were performed on this set of eight tokens, comprising three confusion-inducing and five non-confused tokens.

\begin{figure}[!t]    
    \centering
    \begin{subfigure}{0.45\textwidth} 
        \centering
        \includegraphics[width=\textwidth]{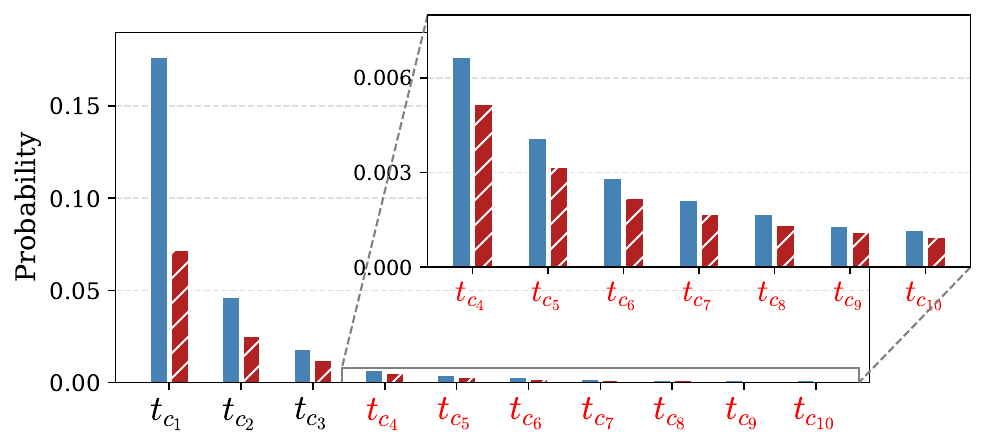}
        \caption{Probability changes of confusion-inducing tokens in cases where three such tokens are included in the top-$N$ candidate set (averaged over 100 samples).
        Solid blue and hatched red bars denote the probabilities before and after the policy update, respectively.}
        \label{fig:fig_probability_belowN_a}
    \end{subfigure}
    
    \vspace{10pt}
    
    \begin{subfigure}{0.48\textwidth}
        \centering
        \includegraphics[width=\textwidth]{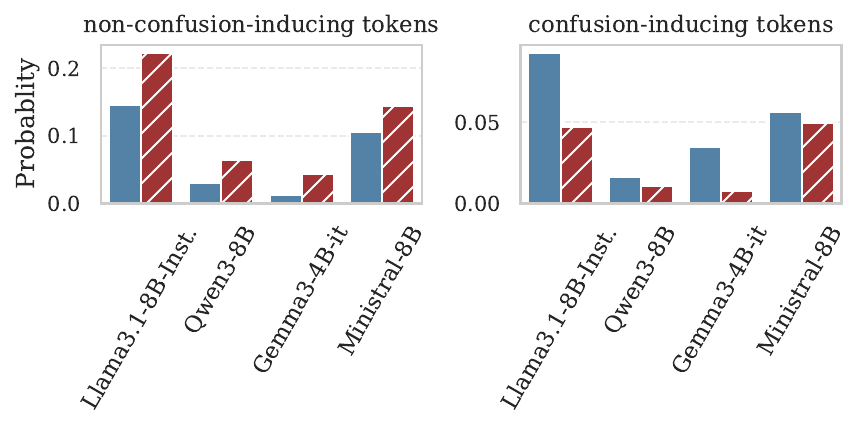}
        \caption{Changes in the cumulative probability of tokens outside the top-$N$ set (i.e., receiving no training signals), separated into confusion-inducing and non-confusion-inducing groups. Solid blue and hatched red bars indicate the values before and after fine-tuning, respectively.}
        \label{fig:fig_probability_belowN_b}        
    \end{subfigure}
    \caption{The impact of TLPO on the probability distributions of tokens outside the top-$N$ set (implicitly affected tokens). 
    }
    \label{fig:fig_probability_belowN}
    \vspace{-5pt}    
\end{figure}

Figure~\ref{fig:fig_probability_belowN}(\subref{fig:fig_probability_belowN_a}) illustrates the changes in output probabilities for confusion-inducing tokens before and after the parameter update. It is observed that the probabilities decrease not only for the tokens explicitly used in the optimization ($t_{c_1}$, $t_{c_2}$, $t_{c_3}$) but also for the remaining confusion-inducing tokens ($t_{c_4},\dots,t_{c_{10}}$) that were not included in the optimization objective.

Furthermore, we extended this analysis to the models evaluated in Section~\ref{sec:main_result} (with $N=16$) to investigate probability shifts for tokens ranked outside the top-$N$. 
Figure~\ref{fig:fig_probability_belowN}(\subref{fig:fig_probability_belowN_b}) illustrates the changes in cumulative probability for tokens outside the top-$N$ set, distinguishing between non-confusion-inducing (Left) and confusion-inducing tokens (Right).
The results demonstrate that under TLPO, the aggregated probability of non-confusion-inducing tokens increases, whereas that of confusion-inducing tokens decreases, for those not explicitly included in the top-$N$ candidate set during fine-tuning. This indicates that the optimization effects of TLPO generalize to tokens that were not explicitly included in the optimization process.

\subsection{Ablation Study on Token Selection and Advantage Formulation}

\begin{figure}[t]
\centering
\includegraphics[width=0.45\textwidth]
{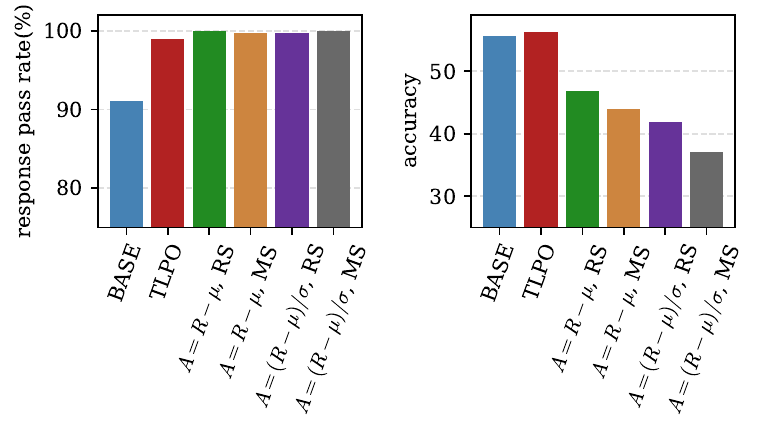}
\caption{Performance Comparison across Different Advantage and Token Selection Strategies. 
In the x-axis labels, RS and MS denote ranked token selection and multinomial sampling respectively.
Additionally, $R$, $\mu$ and $\sigma$ represent the reward, mean reward and standard deviation of the reward. These results were obtained using the Llama-3.1-8B-Instruct model with Korean as the target language.}
\label{fig:fig_ab}
\end{figure}

We conducted an ablation study to evaluate alternative candidate token selection strategies and advantage calculation methods against TLPO. All comparisons were performed within the same framework, where candidate tokens are selected and losses are computed specifically at the confusion point.

For advantage calculation, we compared two variants against our proposed method: the unweighted formulation $A=(R-\mu)$, which excludes the token probability weight and the normalization factor from Eq.~\ref{eq:def_A}, and the GRPO-style advantage defined as $A=(R-\mu)/\sigma$. 
Regarding token selection, we compared our probability-ranked strategy with multinomial sampling.

Figure~\ref{fig:fig_ab} presents the experimental results. The response pass rate remained consistently stable, exceeding 99\% across all settings. However, accuracy exhibited significant variation depending on the method; notably, ranked selection demonstrated superior performance compared to multinomial sampling. Furthermore, in terms of advantage calculation, the specific probability-weighted formulation employed in TLPO achieved the highest performance.
We also observed that the unweighted form $A=(R-\mu)$ outperformed the original GRPO form $A=(R-\mu)/\sigma$. This finding—that excluding the standard deviation normalization yields better performance—aligns with results reported in prior research~\cite{liu2025understanding_r1_zero_like_training}.

\section{Conclusion}

In this paper, we presented Token-Level Policy Optimization (TLPO), a fine-tuning framework designed to mitigate erroneous outputs in multilingual LLMs. Unlike sequence-level methods that optimize entire responses, TLPO operates with precision by updating the policy strictly at specific error positions, thereby resolving inconsistencies without compromising the model's general capabilities. Ultimately, this study offers a new perspective on correcting generative errors: viewing it not as a task of global sequence alignment, but as one of precise, localized adjustment. This approach suggests a promising path forward for fine-grained model alignment beyond language confusion.

\section*{Limitations}

TLPO operates by identifying and rectifying local errors at specific token positions. Consequently, it is particularly effective for tasks where error boundaries are clearly defined, such as mitigating language confusion. However, extending TLPO to tasks relying on holistic sequence-level evaluations, such as general correctness or helpfulness, presents a challenge. In such contexts, errors are often diffuse rather than localized, making it difficult to derive the fine-grained supervision signals necessary for pinpointing and modifying specific tokens.

\paragraph{Broader Impact}
TLPO focuses on correcting linguistic inconsistencies at the token level and does not inherently address the safety or factual validity of the generated content. Consequently, biases or toxic patterns present in the base model or the fine-tuning datasets may be preserved or even amplified in the target language. Users of this framework should ensure that adequate safety filtering and content evaluation measures are in place to mitigate these risks.

\paragraph{AI Assistance Disclosure}
Google Gemini-3 and OpenAI GPT-5 were employed exclusively for refining the clarity and readability of this manuscript.

\bibliography{custom}

@article{schulman2017proximal,
  title={Proximal policy optimization algorithms},
  author={Schulman, John and Wolski, Filip and Dhariwal, Prafulla and Radford, Alec and Klimov, Oleg},
  journal={arXiv preprint arXiv:1707.06347},
  year={2017}
}

@inproceedings{rafailov2023dpo,
  title     = {Direct Preference Optimization: Your Language Model is Secretly a Reward Model},
  author    = {Rafailov, Rafael and Sharma, Archit and Mitchell, Eric and Ermon, Stefano and Manning, Christopher D. and Finn, Chelsea},
  booktitle = {Advances in Neural Information Processing Systems},
  year      = {2023}
}

@inproceedings{hong2024orpo,
  title     = {ORPO: Monolithic Preference Optimization without Reference Model},
  author    = {Hong, Jiwoo and Lee, Noah and Thorne, James},
  booktitle = {Proceedings of the 2024 Conference on Empirical Methods in Natural Language Processing},
  year      = {2024},
  publisher = {Association for Computational Linguistics}
}

@inproceedings{marchisio2024understanding,
  title     = {Understanding and Mitigating Language Confusion in LLMs},
  author    = {Marchisio, Kelly and Ko, Wei-Yin and B{\'e}rard, Alexandre and Dehaze, Th{\'e}o and Ruder, Sebastian},
  booktitle = {Proceedings of the 2024 Conference on Empirical Methods in Natural Language Processing},
  year      = {2024},
  pages     = {6653--6677},
  publisher = {Association for Computational Linguistics}
}

@inproceddings{nie2025mechanistic,
  title   = {Mechanistic Understanding and Mitigation of Language Confusion in English-Centric Large Language Models},
  author  = {Nie, Ercong and Schmid, Helmut and Sch{\"u}tze, Hinrich},
  booktitle = {Findings of the Association for Computational Linguistics: EMNLP 2025},
  year      = {2025},
  pages     = {690--706},
  publisher = {Association for Computational Linguistics}
}

@inproceedings{lee2025controlling,
  title     = {Controlling Language Confusion in Multilingual LLMs},
  author    = {Lee, Nahyun and Woo, Yeongseo and Ko, Hyunwoo and Son, Guijin},
  booktitle = {Proceedings of the 63rd Annual Meeting of the Association for Computational Linguistics: Student Research Workshop},
  year      = {2025},
  publisher = {Association for Computational Linguistics}
}

@inproceedings{oh2025evaluating,
  title     = {Evaluating LLMs' Language Confusion in Code-switching Context},
  author    = {Oh, Juhyun and Yoo, Haneul and Oh, Alice},
  booktitle = {NeurIPS 2025 Workshop on Evaluating the Evolving LLM Lifecycle},
  year      = {2025}
}

@article{lin2025languageconfusiongate,
  title   = {Language Confusion Gate: Language-Aware Decoding Through Model Self-Distillation},
  author={Zhang, Collin and Huang, Fei and Yuan, Chenhan and Lin, Junyang},
  journal = {arXiv preprint arXiv:2510.17555},
  year    = {2025}
}

@article{bloom,
  title   = {BLOOM: A 176B-Parameter Open-Access Multilingual Language Model},
  author  = {Le Scao, Teven and Fan, Angela and Akiki, Christopher and et al.},
  journal = {arXiv preprint arXiv:2211.05100},
  year    = {2022}
}

@article{grattafiori2024llama3,
  title   = {The Llama 3 Herd of Models},
  author  = {Grattafiori, Aaron and et al.},
  journal = {arXiv preprint arXiv:2407.21783},
  year    = {2024}
}

@misc{meta2025llama4,
  title        = {The Llama 4 Herd: The Beginning of a New Era of Natively Multimodal Intelligence},
  author       = {{Meta AI}},
  howpublished = {\url{https://ai.meta.com/blog/llama-4-multimodal-intelligence/}},
  year         = {2025},
  note         = {Meta AI Blog}
}

@article{jiang2024mixtralexperts,
  title        = {Mixtral of Experts},
  author       = {Jiang, Albert Q. and Sablayrolles, Alexandre and Roux, Antoine and Mensch, Arthur and Savary, Blanche and Bamford, Chris and Chaplot, Devendra Singh and de las Casas, Diego and Bou Hanna, Emma and Bressand, Florian and Lengyel, Gianna and Bour, Guillaume and Lample, Guillaume and Lavaud, L{\'e}lio Renard and Saulnier, Lucile and Lachaux, Marie-Anne and Stock, Pierre and Subramanian, Sandeep and Yang, Sophia and Antoniak, Szymon and Le Scao, Teven and Gervet, Th{\'e}ophile and Lavril, Thibaut and Wang, Thomas and Lacroix, Timoth{\'e}e and El Sayed, William},
  journal={arXiv preprint arXiv:2401.04088},
  year={2024}
}

@inproceedings{aya_model,
  title     = {Aya Model: An Instruction Finetuned Open-Access Multilingual Language Model},
  author    = {{\"U}st{\"u}n, Ahmet and Aryabumi, Viraat and Yong, Zheng-Xin and Ko, Wei-Yin and D'souza, Daniel and Onilude, Gbemileke and Bhandari, Neel and Singh, Shivalika and Ooi, Hui-Lee and Kayid, Amr and Vargus, Freddie and Blunsom, Phil and Longpre, Shayne and Muennighoff, Niklas and Fadaee, Marzieh and Kreutzer, Julia and Hooker, Sara},
  booktitle = {Proceedings of the 62nd Annual Meeting of the Association for Computational Linguistics (Volume 1: Long Papers)},
  year      = {2024},
  pages     = {15894--15939},
  address   = {Bangkok, Thailand},
  publisher = {Association for Computational Linguistics}
}

@article{yang2025qwen3,
  title={Qwen3 technical report},
  author={Yang, An and Li, Anfeng and Yang, Baosong and Zhang, Beichen and Hui, Binyuan and Zheng, Bo and Yu, Bowen and Gao, Chang and Huang, Chengen and Lv, Chenxu and others},
  journal={arXiv preprint arXiv:2505.09388},
  year={2025}
}

@article{team2025gemma,
  title={Gemma 3 technical report},
  author={Team, Gemma and Kamath, Aishwarya and Ferret, Johan and Pathak, Shreya and Vieillard, Nino and Merhej, Ramona and Perrin, Sarah and Matejovicova, Tatiana and Ram{\'e}, Alexandre and Rivi{\`e}re, Morgane and others},
  journal={arXiv preprint arXiv:2503.19786},
  year={2025}
}

@article{shao2024deepseekmath,
  title={Deepseekmath: Pushing the limits of mathematical reasoning in open language models},
  author={Shao, Zhihong and Wang, Peiyi and Zhu, Qihao and Xu, Runxin and Song, Junxiao and Bi, Xiao and Zhang, Haowei and Zhang, Mingchuan and Li, YK and Wu, Yang and others},
  journal={arXiv preprint arXiv:2402.03300},
  year={2024}
}

@misc{schulman2020kld,
  title={Approximating KL Divergence},
  author={Schulman, John},
  url={http://joschu.net/blog/kl-approx.html},
  year={2020}
}

@inproceedings{christiano2017deep,
  title     = {Deep Reinforcement Learning from Human Preferences},
  author    = {Christiano, Paul and Leike, Jan and Brown, Tom and Martic, Miljan and Legg, Shane and Amodei, Dario},
  booktitle = {Advances in Neural Information Processing Systems},
  volume    = {30},
  year      = {2017}
}

@inproceedings{ouyang2022training,
  title     = {Training Language Models to Follow Instructions with Human Feedback},
  author    = {Ouyang, Long and Wu, Jeff and Jiang, Xu and Almeida, Diogo and Wainwright, Carroll L. and Mishkin, Pamela and Zhang, Chong and Agarwal, Sandhini and Slama, Katarina and Ray, Alex and Schulman, John and Hilton, Jacob and Kelton, Fraser and Miller, Luke and Simens, Maddie and Askell, Amanda and Welinder, Peter and Christiano, Paul and Leike, Jan and Lowe, Ryan},
  booktitle = {Advances in Neural Information Processing Systems},
  volume    = {35},
  pages     = {27730--27744},
  year      = {2022}
}

@article{ethayarajh2024kto,
  title   = {KTO: Model Alignment as Prospect Theoretic Optimization},
  author  = {Ethayarajh, Kawin and others},
  journal = {arXiv preprint arXiv:2402.01306},
  year    = {2024}
}

@article{kirkpatrick2017overcoming,
  title={Overcoming catastrophic forgetting in neural networks},
  author={Kirkpatrick, James and Pascanu, Razvan and Rabinowitz, Neil and Veness, Joel and Desjardins, Guillaume and Rusu, Andrei A and Milan, Kieran and Quan, John and Ramalho, Tiago and Grabska-Barwinska, Agnieszka and others},
  journal={Proceedings of the national academy of sciences},
  volume={114},
  number={13},
  pages={3521--3526},
  year={2017},
  publisher={National Acad Sciences}
}

@article{luo2023empirical,
  title={An Empirical Study of Catastrophic Forgetting in Large Language Models During Instruction Tuning},
  author={Luo, Yun and Yang, Zhen and Meng, Fandong and Li, Yanan and Zhou, Jie and Zhang, Yue},
  journal={IEEE Transactions on Audio, Speech and Language Processing}, 
  year={2025},
  volume={33},
  number={},
  pages={3776-3786}
}

@inproceedings{lightman2023lets,
  title={Let's Verify Step by Step},
  author={Lightman, Hunter and Kosaraju, Vineet and Burda, Yuri and Edwards, Harrison and Baker, Bowen and Lee, Teddy and Leike, Jan and Schulman, John and Sutskever, Ilya and Cobbe, Karl},
  booktitle={The Twelfth International Conference on Learning Representations},
  year={2024}
}

@inproceedings{conneau2020unsupervised,
  title={Unsupervised Cross-lingual Representation Learning at Scale},
  author={Conneau, Alexis and Khandelwal, Kartikay and Goyal, Naman and Chaudhary, Vishrav and Wenzek, Guillaume and Guzm{\'a}n, Francisco and Grave, Edouard and Ott, Myle and Zettlemoyer, Luke and Stoyanov, Veselin},
  booktitle={Proceedings of the 58th Annual Meeting of the Association for Computational Linguistics},
  pages={8440--8451},
  year={2020}
}

@article{openai2023gpt4,
  title={Gpt-4 technical report},
  author={Achiam, Josh and Adler, Steven and Agarwal, Sandhini and Ahmad, Lama and Akkaya, Ilge and Aleman, Florencia Leoni and Almeida, Diogo and Altenschmidt, Janko and Altman, Sam and Anadkat, Shyamal and others},
  journal={arXiv preprint arXiv:2303.08774},
  year={2023}
}

@article{zhao2023survey,
  title   = {A Survey of Large Language Models},
  author  = {Zhao, Wayne Xin and Zhou, Kun and Li, Junyi and Tang, Tianyi
             and Wang, Xiaolei and Hou, Yupeng and Min, Yingqian and
             Zhang, Beichen and Zhang, Junjie and Dong, Zican and
             Du, Yifan and others},
  journal = {arXiv preprint arXiv:2303.18223},
  year    = {2023}
}

@inproceedings{hu2020xtreme,
  title     = {XTREME: A Massively Multilingual Multi-task Benchmark
               for Evaluating Cross-lingual Generalization},
  author    = {Hu, J. and Ruder, S. and Siddhant, A. and Neubig, G.
               and Firat, O. and Johnson, M.},
  booktitle = {Proceedings of the 37th International Conference on
               Machine Learning},
  year      = {2020}
}

@inproceedings{xu2024finegrained,
  title     = {Aligning Large Language Models via Fine-grained Supervision},
  author    = {Xu, Dehong and Qiu, Liang and Kim, Minseok and
               Ladhak, Faisal and Do, Jaeyoung},
  booktitle = {Proceedings of the 62nd Annual Meeting of the
               Association for Computational Linguistics (ACL)},
  year      = {2024}
}

@inproceedings{zhang2025tokenlevelacceptrejectmicro,
  title     = {Token-level Accept or Reject: A Micro Alignment Approach
               for Large Language Models},
  author    = {Zhang, Yang and Yu, Yu and Tang, Bo and Zhu, Yu and
               Sun, Chuxiong and Wei, Wenqiang and Hu, Jie and
               Xie, Zipeng and Li, Zhiyu and Xiong, Feiyu and
               Chung, Edward},
  booktitle = {Proceedings of the 34th International Joint Conference
               on Artificial Intelligence (IJCAI)},
  year      = {2025}
}

@misc{li2023bactrianx,
  title         = {Bactrian-X: A Multilingual Replicable Instruction-Following Model with Low-Rank Adaptation},
  author        = {Haonan Li and Fajri Koto and Minghao Wu and Alham Fikri Aji and Timothy Baldwin},
  year          = {2023},
  eprint        = {2305.15011},
  archivePrefix = {arXiv},
  primaryClass  = {cs.CL}
}

@inproceedings{zeng2024marcobenchmif,
  title={Marco-Bench-MIF: On Multilingual Instruction-Following Capability of Large Language},
  author={Zeng, Bo and Lyu, Chenyang and Liu, Sinuo and Zeng, Mingyan and Wu, Minghao and Ni, Xuanfan and Shi, Tianqi and Zhao, Yu and Liu, Yefeng and Zhu, Chenyu and others},
  booktitle={Proceedings of the 63rd Annual Meeting of the Association for Computational Linguistics (Volume 1: Long Papers)},
  pages={24058--24072},
  year={2025}
}

@misc{openai2024mmmlu,
  title        = {Multilingual Massive Multitask Language Understanding (MMMLU)},
  author       = {{OpenAI}},
  year         = {2024},
  howpublished = {\url{https://huggingface.co/datasets/openai/MMMLU}},
  note         = {Dataset}
}

@article{hendryckstest2021,
  title   = {Measuring Massive Multitask Language Understanding},
  author  = {Dan Hendrycks and Collin Burns and Steven Basart and Andy Zou and
             Mantas Mazeika and Dawn Song and Jacob Steinhardt},
  journal = {Proceedings of the International Conference on Learning Representations},
  year    = {2021}
}

@inproceedings{rein2023gpqa,
  title  = {{GPQA}: A Graduate-Level Google-Proof Q\&A Benchmark},
  author={Rein, David and Hou, Betty Li and Stickland, Asa Cooper and Petty, Jackson and Pang, Richard Yuanzhe and Dirani, Julien and Michael, Julian and Bowman, Samuel R},
  booktitle={First Conference on Language Modeling},
  year={2024}
}

@article{clark2018arc,
  title   = {Think you have Solved Question Answering? Try {ARC}, the {AI2} Reasoning Challenge},
  author  = {Peter Clark and Isaac Cowhey and Oren Etzioni and Tushar Khot and
             Ashish Sabharwal and Carissa Schoenick and Oyvind Tafjord},
  journal = {arXiv preprint arXiv:1803.05457},
  year    = {2018}
}

@inproceedings{suzgun2022challenging,
  title     = {Challenging Big-Bench Tasks and Whether Chain-of-Thought Can Solve Them},
  author={Suzgun, Mirac and Scales, Nathan and Sch{\"a}rli, Nathanael and Gehrmann, Sebastian and Tay, Yi and Chung, Hyung Won and Chowdhery, Aakanksha and Le, Quoc and Chi, Ed and Zhou, Denny and others},
  booktitle={Findings of the Association for Computational Linguistics: ACL 2023},
  pages={13003--13051},
  year={2023}
}

@article{cobbe2021gsm8k,
  title   = {Training verifiers to solve math word problems},
  author  = {Karl Cobbe and Vineet Kosaraju and Mohammad Bavarian and Mark Chen and Lukasz Kaiser and Matthias Plappert and Jerry Tworek and Jacob Hilton and  Reiichiro Nakano and Christopher Hesse and John Schulman},
  journal = {arXiv preprint arXiv:2110.14168},
  year    = {2021}
}

@article{hendrycks2021math,
  title   = {Measuring Mathematical Problem Solving With the MATH Dataset},
  author  = {Dan Hendrycks and Collin Burns and Saurav Kadavath and Akul Arora and
             Steven Basart and Eric Tang and Dawn Song and Jacob Steinhardt},
  journal = {arXiv preprint arXiv:2103.03874},
  year    = {2021}
}

@inproceedings{
liu2025understanding_r1_zero_like_training,
title={Understanding R1-Zero-Like Training: A Critical Perspective},
author={Zichen Liu and Changyu Chen and Wenjun Li and Penghui Qi and Tianyu Pang and Chao Du and Wee Sun Lee and Min Lin},
booktitle={Second Conference on Language Modeling},
year={2025},
url={https://openreview.net/forum?id=5PAF7PAY2Y}
}

@misc{guo2025deepseek,
      title={DeepSeek-R1: Incentivizing Reasoning Capability in LLMs via Reinforcement Learning}, 
      author={Daya Guo and Dejian Yang and Haowei Zhang and Junxiao Song and Peiyi Wang and Qihao Zhu and Runxin Xu and Ruoyu Zhang and Shirong Ma and others},
      year={2025},
      eprint={2501.12948},
      archivePrefix={arXiv},
      primaryClass={cs.CL}
}

\appendix

\section{Algorithm of TLPO}
\label{sec:appendix_tlpo_algorithm}

\begin{algorithm}[!h]
\fontsize{9.5pt}{11pt}\selectfont
\caption{Token Level Policy Optimization (TLPO)}
\algrenewcommand\algorithmicindent{1.0em}
\begin{algorithmic}[1]

    \Require initial policy model $\pi_{\theta_{\mathrm{ref}}}$, 
            prompt dataset $\mathcal{D}$,
            learning rate $\alpha$,
            Top-N size $N$,
            training steps $M$,
            TLPO iterations $p$
    
    \State Initialize the target parameters: $\theta \gets \theta_{\mathrm{ref}}$   
    
    \For {step=1,...,$M$}
    \State \text{Sample an input prompt batch $X$ from $\mathcal{D}$} 

    \For {$x \in X$}
        \State Sample an output sequence $y$ from $\pi_{\theta}(\cdot \mid x)$
        \State Detect the confusion point $c$ in $y$
        \State Set $\mathcal{T} = \{t_i\}_{i=1}^N$ as $\operatorname*{topN}$ tokens of $\pi_{\theta}(\cdot|x,y_{<c})$
         
        \State Obtain reward values $r_i \gets R(t_i)$, $\forall t_i \in \mathcal{T}$  
    \EndFor

    \State  $\theta_{\mathrm{old}} \gets \theta$

    \For {TLPO iteration=1,...,$p$}
        \State Compute the objective ${J_{\mathrm{TLPO}}(\theta)}$ via Eq.~(\ref{eq:tlpo_j})

        \State Compute gradient $g \leftarrow \nabla_{\theta} {J}_{\mathrm{TLPO}}(\theta)$

        \State Update policy parameters: $\theta \gets \theta + \alpha \cdot g$
    
    \EndFor

    \EndFor

    \Ensure optimized policy $\pi_{\theta}$
\end{algorithmic}
\label{algo:tlpo}
\end{algorithm}

Algorithm~\ref{algo:tlpo} outlines the overall procedure of Token-Level Policy Optimization (TLPO). The TLPO algorithm comprises two primary phases: an \textbf{exploration phase}, which detects a confusion point and selects candidate tokens at that position, and a \textbf{policy update phase}, which optimizes the policy using these selected candidates.

\paragraph{Exploration (Lines 3--9)}
First, a batch of prompts $X$ is sampled from the training dataset $\mathcal{D}$. For each prompt $x \in X$, the model generates an initial response $y$. The algorithm then detects a confusion point $c$ within $y$. If no confusion is detected, the corresponding sample is discarded. Conversely, if a confusion point is identified, the top-$N$ tokens are selected from the current policy distribution $\pi_{\theta}(\cdot|x, y_{<c})$ to form the candidate token set $\mathcal{T}$. Subsequently, a reward is assigned to each candidate token based on a short lookahead rollout of length $k$.

\paragraph{Policy Update (Lines 10--15)}
Once rewards for all candidate tokens in the batch are collected, the policy parameters $\theta$ are updated. Specifically, we perform $p$ optimization iterations to maximize the TLPO objective function $\mathcal{J}_{\mathrm{TLPO}}(\theta)$, following a methodology similar to Proximal Policy Optimization (PPO).

\section{Training Dataset Construction}
\label{sec:appendix_train_dataset}

\begin{table}[!tb]
\centering
\fontsize{10pt}{11pt}\selectfont

\begin{tabular}{c|c|c}
\toprule
\multirow{2}{*}{Language} & Original training & Filtered training \\
& {\#instances} & {\#instances} \\

\midrule

Chinese(zh)    & 67,017 & 65,676  \\
Arabic(ar)    & 67,017 & 65,907  \\
Korean(ko)    & 67,017 & 62,679 \\
Japanese(ja)    & 67,017 & 65,296 \\

\bottomrule
\end{tabular}

\caption{Number of training instances.}
\label{tab:training_data}
\end{table}

To construct the fine-tuning dataset for mitigating language confusion, we utilized the training split of the Bactrian-X dataset for each target language. Prompts were formed by concatenating the \textit{instruction} and \textit{input} fields from the dataset, while the \textit{output} field served as the answer.

To exclude prompts that explicitly induce generation in other languages (e.g., translation requests), 
we filtered out any prompts containing characters that do not belong to the target language. 
Table~\ref{tab:training_data} presents the number of prompts in the original dataset and the final number of prompts used for training after this filtering process.

The specific configurations for each fine-tuning method are as follows:

\paragraph{SFT} We conducted Supervised Fine-Tuning (SFT) using the filtered prompts and their corresponding answers from the Bactrian-X dataset for each target language.

\paragraph{DPO and ORPO} For these methods, we generated 16 candidate responses for each prompt using the respective target models. We then constructed preference pairs for fine-tuning by selecting one response without \textit{language confusion} as the \textbf{preferred} response and one exhibiting confusion as the \textbf{dispreferred} response.

\paragraph{TLPO}
For TLPO, fine-tuning was performed by generating responses online based on the prompts from the training set. Specifically, we generated a single full response per prompt.

\section{Training Configurations}

TLPO fine-tuning was performed using a unified hyperparameter configuration across all models and target languages. The specific settings are as follows:
\begin{itemize}

\item \textbf{General Training Settings:} All experiments were conducted for 1 epoch. The batch size was set to 8, representing the number of prompts processed in a single step. However, since the loss is computed over 16 candidate tokens ($N=16$) at each confusion point, the parameters are updated based on a total of $8 \times 16 = 128$ tokens. We used an initial learning rate of $5 \times 10^{-7}$ with a cosine decay schedule, reducing the rate to 10\% of the initial value by the end of training. A warmup period covering the first 10\% of total steps was applied.

\item \textbf{TLPO-Specific Parameters:} We set the number of policy iterations to 2 ($p=2$). At each language confusion point, we explored 16 candidate tokens ($N=16$) and employed a lookahead length of 3 tokens ($k=3$) for reward calculation.
\end{itemize}


\section{Computational Cost and Training Time}
\label{sec:appendix_training_time}

\begin{table}[t]
\centering
\fontsize{9.5pt}{11pt}\selectfont

\begin{tabular}{c|c|c|c|c}
\toprule

& \multicolumn{4}{c}{Fine-tuning Time (hours)} \\ 
\cmidrule(lr){2-5}
\multirow{2}{*}{Lang.}
& Llama3.1 & Qwen3 & Ministral & Gemma3 \\
& -8B-Inst. & -8B & -8B & -4B-it \\

\midrule
\midrule

zh   & 6.81 & 7.95 & 7.04 & 6.80 \\
ar   & 6.20 & 7.81 & 7.91 & 5.78 \\
ko   & 8.27 & 8.97 & 7.95 & 6.26 \\
ja   & 7.05 & 9.59 & 7.27 & 8.83 \\

\bottomrule
\end{tabular}
\caption{TLPO fine-tuning time for each model and target language. All experiments were conducted using 8x NVIDIA H100 GPUs.}
\label{tab:app_training_time_detail}
\end{table}

Table~\ref{tab:app_training_time_detail} presents the wall-clock time required for TLPO fine-tuning across different model and language configurations. Fine-tuning was performed for a single epoch, and the results indicate that training is completed within approximately 6 to 10 hours depending on the specific model and language pair.


\begin{table}[t]
\centering
\fontsize{8.5pt}{11pt}\selectfont

\begin{tabular}{c|c|c|c}
\toprule
\multirow{2}{*}{Task} & WPR & \multirow{2}{*}{Acc.} & \multirow{2}{*}{\#Instances} \\
& /RPR & & \\
\midrule
\midrule

LCB(cross-lingual)    & O &   & 299 \\
LCB(monolingual)      & O &   & 200/300/100/100 \\
MIF(target lang.)     & O & O & 420/421/422/421 \\
MMMLU(target lang.)   & O & O & 14,042 \\
GSM8K(cross)          & O & O & 1,209  \\

\midrule

MIF(en)               &  & O & 541 \\
GPQA(en)              &  & O & 448 \\
GPQA(diamond, en)     &  & O & 198 \\
ARC-C(en)             &  & O & 1,172 \\
BBH(en)               &  & O & 6,511 \\
MATH(en)              &  & O & 5,000 \\
GSM8K(en)             &  & O & 1,209\\

\bottomrule
\end{tabular}

\caption{List of tasks used for consistency (WPR/RPR) and accuracy evaluation. For LCB (monolingual) and MIF (target), the values for \#instances are presented in the order of zh/ar/ko/ja.}
\label{tab:app_task_detail}
\end{table}

\section{Evaluation Methodology}
\label{sec:appendix_eval_dataset}

Table~\ref{tab:app_task_detail} presents the tasks used for WPR, RPR, and Accuracy evaluation, along with the number of instances for each task.
All evaluations were conducted as generative tasks under a zero-shot Chain-of-Thought (CoT) setting.

For the consistency evaluation on the MIF dataset, we excluded specific instances where the instruction explicitly requires generating output in a different language, ensuring that the metric accurately reflects unintended language confusion.

\begin{figure}[!htbp]
\begin{center}
\includegraphics[width=0.98\columnwidth]{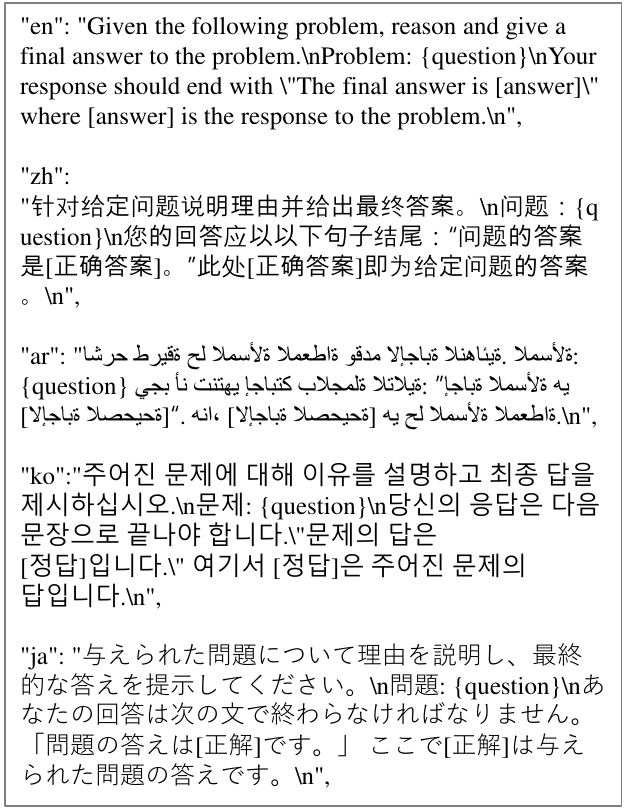}
\vspace{-5pt}
\caption{
Instruction used for GSM8K(en) and GSM8K(cross). Note that \{question\} represents the original English question from the GSM8K dataset, which remains untranslated.
}
\label{fig:gsm8k_instructions}
\end{center}
\end{figure}

In the case of GSM8K (cross), we adapted the prompt template from lm-eval-harness (specifically gsm8k-cot-llama). The English instructions were translated into each target language to serve as the prompts. Figure~\ref{fig:gsm8k_instructions} illustrates the specific instructions used for each target language.

\section{Details of the Language Confusion Detector}
\label{sec:appendix_detect_lc_detail}

We devised a rule-based heuristic to detect language confusion. This detector operates by analyzing an LLM-generated response to count the number of words exhibiting language confusion versus those that do not. Based on these counts, we subsequently calculate the Word Pass Rate (WPR) and Response Pass Rate (RPR).

For word segmentation, we utilized the \texttt{jieba} library for Chinese and a Python-based tagger library for Japanese. For all other languages, segmentation was performed based on whitespace characters.

We determined whether each character within a word belongs to the target language by referencing its Unicode metadata (e.g., script/block information derived from character names). A word was classified as free of language confusion only if all its constituent characters belonged to the target language. Conversely, if a word contained one or more characters not belonging to the target language, it was classified as exhibiting language confusion.

During the detection process, we applied several exclusion rules to prevent false detections:
\begin{itemize}
\item \textbf{Word/Line-level exclusions:} Units of measurement denoted in the alphabet, strings identified as function names, email addresses, and URLs were excluded from detection.
\item \textbf{Character-level exclusions:} We also excluded phonetic symbols, words starting with a capital letter (indicating proper nouns), mathematical symbols, currency symbols, arrows, Chinese tone marks, and emojis.
\end{itemize}

\vspace{0.5em}
To validate the effectiveness of these rules, we conducted a systematic error analysis. 
First, the likelihood of \textbf{False Negatives (FN)} is practically zero due to the deterministic nature of Unicode script mapping; every character belonging to the target script is correctly identified without exception. 
Second, a manual inspection of 10,937 detected instances revealed only 9 \textbf{False Positives (FP)}, yielding a remarkably low error rate of 0.08\%. 
These rare FPs were primarily caused by uncommon emojis not yet included in our exclusion list. 
Overall, this high level of precision ensures that our detector provides a clean and accurate signal for optimizing multilingual alignment.

\section{Performance Sensitivity to the Number of Candidate Tokens}

\begin{figure}[t] 
    \centering   
    \includegraphics[width=0.45\textwidth]{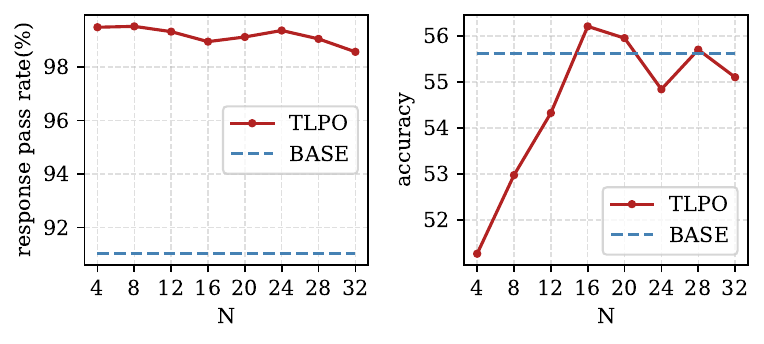} 
    \caption{Response Pass Rate and Accuracy across Various Numbers of Candidate Tokens ($N$). This experiment was conducted using the Llama3.1-8B-Instruct model with Korean as the target language.}
    \label{fig:varN}
    \vspace{-10pt}
\end{figure}

\begin{figure*}[htbp!] 
    \centering   
    \includegraphics[width=0.98\textwidth]{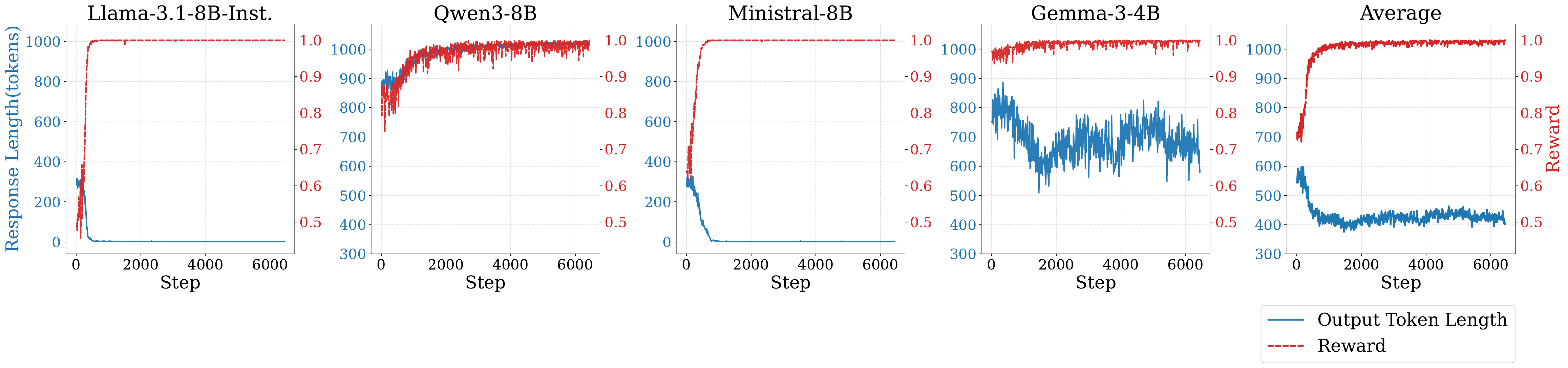} 
    \caption{Response length and reward dynamics during the GRPO fine-tuning process. Blue lines (left axis) and red lines (right axis) represent response length and reward, respectively.}
    \label{fig:tlpo_training_results}
    \vspace{-10pt}
\end{figure*}

Figure~\ref{fig:varN} illustrates the variations in response pass rate and accuracy as the number of candidate tokens $N$ explored at the confusion point $c$ changes. The response pass rate remains consistently high at over 99\% regardless of $N$, showing negligible fluctuation. In contrast, accuracy exhibits a notable decline in the range of $N \le 12$, dropping by 4.4\,pp, 2.6\,pp and 1.29\,pp compared to the Baseline at $N=4, 8$ and $N=12$, respectively. However, for $N \ge 16$, the decrease in accuracy narrows to less than 1\,pp, demonstrating that performance is stably preserved.


\section{Supplementary Results on GRPO Fine-tuning Dynamics}

Figure~\ref{fig:tlpo_training_results} illustrates the dynamics of response length and reward throughout the GRPO fine-tuning process. For Llama-3.1-8B-Instruct and Ministral-8B, a sharp decline in response length was observed immediately upon the start of fine-tuning, while Gemma3-4B exhibited a reduction in output length to approximately three-quarters of its initial size. Conversely, Qwen3-8B was the only model that maintained stable token length without such degradation.

These experiments were conducted under a setting where English occurrences are treated as a neutral category rather than language confusion. In this setup, we assigned a reward of -1 for instances of language confusion (non-target languages excluding English) and +1 for proper linguistic adherence. We attribute the observed length reduction to the model's exploitation of the reward structure; by shortening its responses, the model effectively minimizes the accumulation of negative rewards. Due to this model collapse phenomenon, we excluded GRPO from the primary baselines in our main experimental results.

\section{Detailed Experimental Results by Model and Target Language}
\label{sec:appendix_detail_result}

\begin{figure*}[!tb]
\begin{center}
\includegraphics[width=0.98\linewidth]{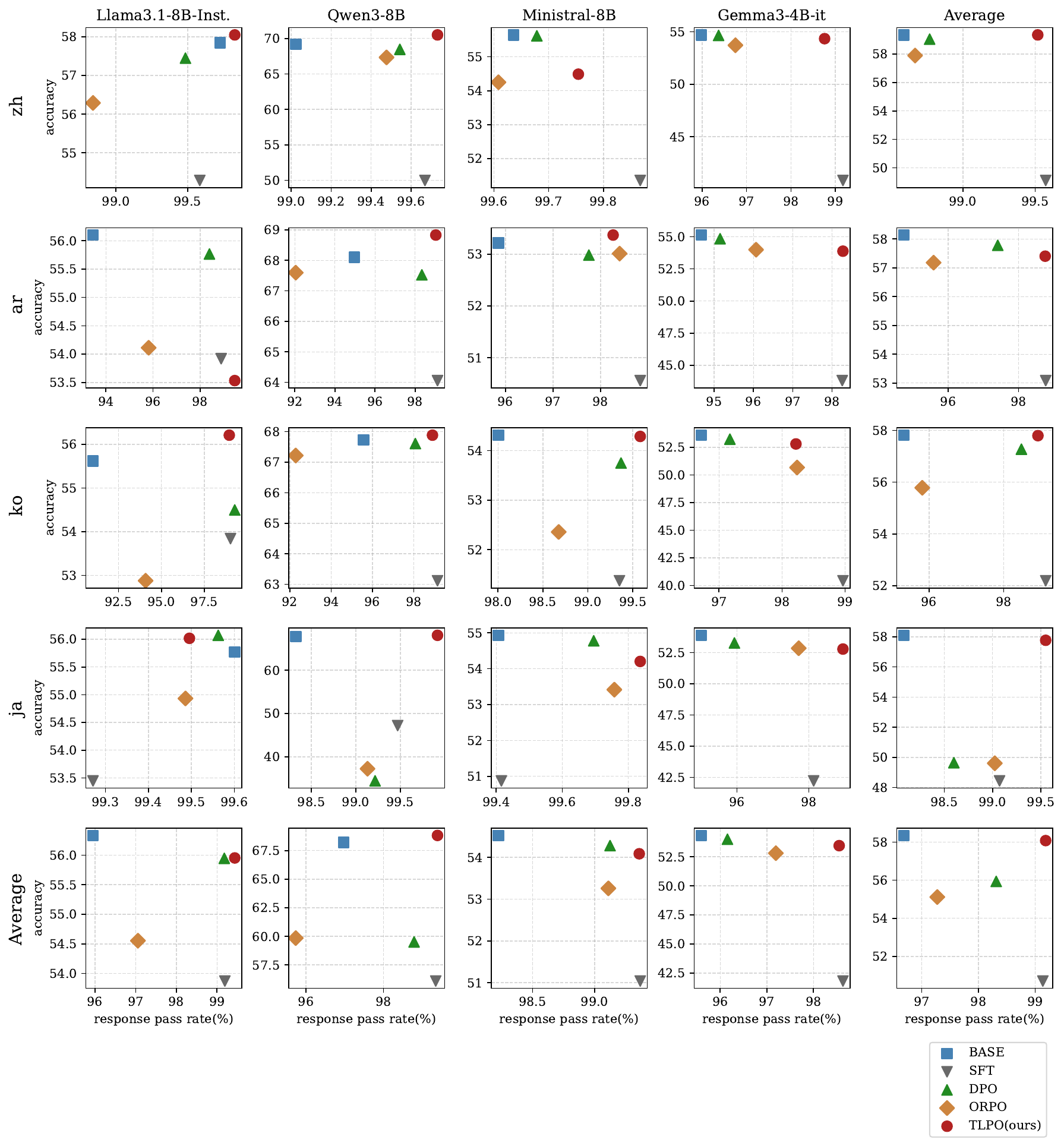}
\caption{Response Pass Rate (RPR) and accuracy plots across models and target languages, under the setting where English is treated as neutral.}
\label{fig:main_result_iet_all}
\end{center}
\end{figure*}

Figure~\ref{fig:main_result_iet_all} presents the Response Pass Rate (RPR) and accuracy for each model across different target languages after fine-tuning to mitigate language confusion, under the setting where English is treated as a neutral language. The results demonstrate that in most configurations, TLPO effectively mitigates language confusion while minimizing the degradation of the LLM's performance more consistently than other comparative methods. Tables~\ref{tab:main_result_llama} through Tables~\ref{tab:main_result_avg} provide detailed RPR, WPR, and accuracy results for each downstream task, broken down by model and target language after fine-tuning.

\begin{table*}[!ht]
    \fontsize{7.6pt}{9pt}\selectfont
    \centering
    
    \begin{subtable}{0.98\textwidth}
    \centering
    \begin{tabular}{c||c||ccccc|c}
    \toprule
    \multirow{2}{*}{Lang.} & \multirow{2}{*}{Method} & LCB & LCB & MIF & MMMLU & GSM8K  & \multirow{2}{*}{Mean} \\
    & & (cross-lingual) & (monolingual) & (target) & (target) & (cross)  & \\
    \midrule 
    
    \midrule     
    \multirow{5}{*}{\shortstack{\textbf{zh}}}  
        & Baseline & 98.96(100.00) & 100.00(100.00) & 100.00(100.00) & 99.76(100.00) & 99.92(100.00) & 99.73(100.00) \\
         & SFT & 98.33(99.99) & 100.00(100.00) & 99.76(100.00) & 99.92(100.00) & 99.92(99.99) & 99.59(100.00) \\
         & DPO & 98.26(99.17) & 100.00(100.00) & 99.52(99.99) & 99.72(99.89) & 99.92(99.82) & 99.48(99.77) \\
         & ORPO & 96.91(99.96) & 98.50(99.98) & 99.29(99.99) & 99.67(99.98) & 99.83(100.00) & 98.84(99.98) \\
         & TLPO(ours) & 99.29(100.00) & 100.00(100.00) & 100.00(100.00) & 99.86(100.00) & 100.00(100.00) & 99.83(100.00) \\
    \midrule 
    \multirow{5}{*}{\shortstack{\textbf{ar}}}  
        & Baseline & 86.21(99.93) & 94.33(99.97) & 95.49(99.97) & 96.69(99.97) & 94.54(99.86) & 93.45(99.94) \\
         & SFT & 96.23(99.98) & 99.33(100.00) & 100.00(100.00) & 99.91(100.00) & 99.00(99.97) & 98.89(99.99) \\
         & DPO & 95.92(99.01) & 99.33(99.99) & 99.29(99.87) & 99.19(99.40) & 98.26(98.16) & 98.40(99.29) \\
         & ORPO & 90.69(99.95) & 95.67(99.97) & 98.57(99.99) & 97.78(99.98) & 96.36(99.80) & 95.82(99.94) \\
         & TLPO(ours) & 97.81(99.97) & 100.00(100.00) & 100.00(100.00) & 99.80(100.00) & 99.75(100.00) & 99.47(99.99) \\

    \midrule 
    \multirow{5}{*}{\shortstack{\textbf{ko}}}
        & Baseline & 87.76(99.81) & 86.00(99.85) & 92.58(99.94) & 94.86(99.93) & 93.88(99.70) & 91.02(99.84) \\
         & SFT & 97.60(99.98) & 100.00(100.00) & 98.52(99.98) & 99.65(99.98) & 99.34(99.93) & 99.02(99.98) \\
         & DPO & 97.56(99.99) & 100.00(100.00) & 99.52(99.99) & 99.59(99.99) & 99.67(99.99) & 99.27(99.99) \\
         & ORPO & 91.10(99.77) & 94.00(99.94) & 94.51(99.89) & 96.00(99.93) & 94.79(99.71) & 94.08(99.85) \\
         & TLPO(ours) & 96.98(99.88) & 100.00(100.00) & 99.04(99.99) & 99.47(99.99) & 99.26(99.95) & 98.95(99.96) \\

    \midrule 
    \multirow{5}{*}{\shortstack{\textbf{ja}}}  
        & Baseline & 98.51(99.99) & 100.00(100.00) & 99.76(100.00) & 99.74(100.00) & 100.00(100.00) & 99.60(100.00) \\
         & SFT & 97.01(99.96) & 100.00(100.00) & 99.51(100.00) & 99.92(100.00) & 99.91(100.00) & 99.27(99.99) \\
         & DPO & 99.59(100.00) & 99.00(100.00) & 99.52(99.98) & 99.78(99.97) & 99.92(99.96) & 99.56(99.98) \\
         & ORPO & 98.13(99.99) & 100.00(100.00) & 99.52(99.95) & 99.77(100.00) & 100.00(100.00) & 99.49(99.99) \\
         & TLPO(ours) & 98.08(99.99) & 100.00(100.00) & 99.52(99.94) & 99.88(100.00) & 100.00(100.00) & 99.50(99.99) \\

    \midrule 
    \multirow{5}{*}{\shortstack{\textbf{avg.}}}  
        & Baseline & 92.86(99.93) & 95.08(99.95) & 96.96(99.98) & 97.76(99.97) & 97.08(99.89) & 95.95(99.94) \\
         & SFT & 97.29(99.98) & 99.83(100.00) & 99.45(99.99) & 99.85(99.99) & 99.54(99.97) & 99.19(99.99) \\
         & DPO & 97.83(99.54) & 99.58(100.00) & 99.46(99.96) & 99.57(99.81) & 99.44(99.48) & 99.18(99.76) \\
         & ORPO & 94.21(99.92) & 97.04(99.97) & 97.97(99.95) & 98.31(99.97) & 97.75(99.88) & 97.06(99.94) \\
         & TLPO(ours) & 98.04(99.96) & 100.00(100.00) & 99.64(99.98) & 99.75(100.00) & 99.75(99.99) & 99.44(99.99) \\

    \bottomrule         
    \end{tabular}

    \caption{Average Response Pass Rate(RPR) and Word Pass Rate(WPR). Values are presented as RPR(WPR) in \%.}        
    \end{subtable}

    \par\bigskip

    \begin{subtable}{0.98\textwidth}
    \centering
    \begin{tabular}{c||c||cccccccccc|c}
    \toprule
         \multirow{2}{*}{\fontsize{7pt}{9pt}\selectfont Lang.} 
         & \multirow{2}{*}{\fontsize{7pt}{9pt}\selectfont Method} 
         & \fontsize{7pt}{9pt}\selectfont MIF
         & \fontsize{7pt}{9pt}\selectfont MIF
         & \fontsize{6.5pt}{9pt}\selectfont MMMLU
         & \fontsize{7pt}{9pt}\selectfont GPQA
         & \fontsize{6.5pt}{9pt}\selectfont GPQA-D
         & \fontsize{6.5pt}{9pt}\selectfont ARC-C
         & \fontsize{7pt}{9pt}\selectfont BBH
         & \fontsize{7pt}{9pt}\selectfont MATH
         & \fontsize{7pt}{9pt}\selectfont GSM8K
         & \fontsize{7pt}{9pt}\selectfont GSM8K
         & \multirow{2}{*}{\fontsize{7pt}{9pt}\selectfont Mean} \\

         &
         & \fontsize{7pt}{9pt}\selectfont (en)       
         & \fontsize{7pt}{9pt}\selectfont (target) 
         & \fontsize{7pt}{9pt}\selectfont (target) 
         & \fontsize{7pt}{9pt}\selectfont (en)  
         & \fontsize{7pt}{9pt}\selectfont (en)
         & \fontsize{7pt}{9pt}\selectfont (en)  
         & \fontsize{7pt}{9pt}\selectfont (en)      
         & \fontsize{7pt}{9pt}\selectfont (en)  
         & \fontsize{7pt}{9pt}\selectfont (en)  
         & \fontsize{7pt}{9pt}\selectfont (cross) 
         &  \\

    \midrule 
    
    \midrule     
    \multirow{5}{*}{\fontsize{7pt}{9pt}\selectfont \shortstack{\textbf{zh}}}     
         & Baseline & 74.68 & 51.57 & 55.32 & 27.46 & 26.26 & 83.79 & 51.27 & 49.22 & 79.24 & 79.65 & 57.85 \\
         & SFT & 71.35 & 48.06 & 41.92 & 25.67 & 22.73 & 84.13 & 56.90 & 43.34 & 76.76 & 71.96 & 54.28 \\
         & DPO & 75.42 & 50.46 & 53.68 & 26.34 & 26.26 & 83.70 & 51.05 & 48.76 & 78.91 & 79.90 & 57.45 \\
         & ORPO & 72.09 & 45.10 & 49.47 & 26.34 & 25.76 & 83.79 & 51.88 & 49.50 & 78.66 & 80.31 & 56.29 \\
         & \fontsize{7pt}{9pt}\selectfont{TLPO(ours)} & 73.57 & 54.90 & 55.73 & 23.88 & 26.77 & 83.02 & 52.79 & 49.56 & 79.90 & 80.40 & 58.05 \\

    \midrule     
    \multirow{5}{*}{\fontsize{7pt}{9pt}\selectfont \shortstack{\textbf{ar}}}     
         & Baseline & 74.31 & 46.03 & 47.19 & 26.34 & 28.28 & 83.62 & 51.70 & 49.78 & 78.91 & 74.86 & 56.10 \\
         & SFT & 71.35 & 47.69 & 36.41 & 28.79 & 25.25 & 83.96 & 58.35 & 44.30 & 77.34 & 65.76 & 53.92 \\
         & DPO & 73.75 & 44.36 & 43.62 & 29.46 & 29.29 & 83.53 & 52.05 & 48.82 & 79.74 & 73.04 & 55.77 \\
         & ORPO & 70.98 & 40.30 & 39.61 & 26.56 & 26.26 & 84.13 & 49.13 & 48.90 & 78.16 & 77.09 & 54.11 \\
         & \fontsize{7pt}{9pt}\selectfont{TLPO(ours)} & 70.61 & 38.08 & 42.62 & 26.34 & 30.30 & 83.70 & 52.10 & 47.08 & 79.32 & 65.18 & 53.53 \\

    \midrule     
    \multirow{5}{*}{\fontsize{7pt}{9pt}\selectfont \shortstack{\textbf{ko}}}     
         & Baseline & 74.86 & 40.85 & 49.15 & 26.79 & 25.25 & 83.70 & 50.98 & 49.16 & 79.57 & 75.85 & 55.62 \\
         & SFT & 73.94 & 41.22 & 34.52 & 27.46 & 29.80 & 83.79 & 57.96 & 43.66 & 76.51 & 69.56 & 53.84 \\
         & DPO & 75.42 & 40.30 & 46.78 & 25.22 & 22.73 & 83.70 & 50.55 & 48.16 & 78.99 & 73.12 & 54.50 \\
         & ORPO & 74.31 & 31.98 & 34.64 & 26.34 & 24.24 & 84.13 & 50.28 & 48.10 & 78.83 & 75.93 & 52.88 \\
         & \fontsize{7pt}{9pt}\selectfont{TLPO(ours)} & 71.90 & 46.40 & 47.51 & 27.23 & 28.28 & 83.62 & 52.00 & 49.14 & 79.40 & 76.59 & 56.21 \\

    \midrule     
    \multirow{5}{*}{\fontsize{7pt}{9pt}\selectfont \shortstack{\textbf{ja}}}     
         & Baseline & 75.05 & 41.59 & 50.44 & 25.00 & 25.25 & 83.70 & 52.00 & 49.74 & 78.00 & 76.92 & 55.77 \\
         & SFT & 73.38 & 41.40 & 37.16 & 29.46 & 25.76 & 83.79 & 57.70 & 43.38 & 76.18 & 66.25 & 53.45 \\
         & DPO & 73.75 & 45.47 & 49.95 & 23.66 & 29.29 & 83.79 & 51.13 & 48.58 & 79.98 & 75.10 & 56.07 \\
         & ORPO & 72.46 & 36.97 & 48.80 & 24.55 & 23.74 & 83.62 & 51.96 & 50.08 & 78.99 & 78.16 & 54.93 \\
         & \fontsize{7pt}{9pt}\selectfont{TLPO(ours)} & 73.75 & 42.70 & 50.11 & 27.01 & 26.77 & 84.04 & 51.28 & 49.00 & 79.16 & 76.34 & 56.02 \\

    \midrule     
    \multirow{5}{*}{\fontsize{7pt}{9pt}\selectfont \shortstack{\textbf{avg.}}}     
         & Baseline & 74.72 & 45.01 & 50.53 & 26.40 & 26.26 & 83.70 & 51.49 & 49.47 & 78.93 & 76.82 & 56.33 \\
         & SFT & 72.50 & 44.59 & 37.50 & 27.85 & 25.88 & 83.92 & 57.73 & 43.67 & 76.70 & 68.38 & 53.87 \\
         & DPO & 74.59 & 45.15 & 48.51 & 26.17 & 26.89 & 83.68 & 51.20 & 48.58 & 79.40 & 75.29 & 55.95 \\
         & ORPO & 72.46 & 38.59 & 43.13 & 25.95 & 25.00 & 83.92 & 50.81 & 49.14 & 78.66 & 77.87 & 54.55 \\
         & \fontsize{7pt}{9pt}\selectfont{TLPO(ours)} & 72.46 & 45.52 & 48.99 & 26.12 & 28.03 & 83.60 & 52.04 & 48.70 & 79.45 & 74.63 & 55.95 \\

    \bottomrule         
    \end{tabular}

    \caption{Average accuracy after fine-tuning.}        
    \end{subtable}

    \caption{
    Detailed RPR, WPR, and accuracy results for the Llama3.1-8B-Instruction model after fine-tuning, in a setting where English output is regarded as neutral.
    }
    
    \label{tab:main_result_llama}
    \vspace{-10pt}    
\end{table*}


\begin{table*}[!ht]
    \fontsize{7.6pt}{9pt}\selectfont
    \centering
    
    \begin{subtable}{0.98\textwidth}
    \centering
    \begin{tabular}{c||c||ccccc|c}
    \toprule
    \multirow{2}{*}{Lang.} & \multirow{2}{*}{Method} & LCB & LCB & MIF & MMMLU & GSM8K  & \multirow{2}{*}{Mean} \\
    & & (cross-lingual) & (monolingual) & (target) & (target) & (cross)  & \\
    \midrule 
    
    \midrule     
    \multirow{5}{*}{\shortstack{\textbf{zh}}}  
         & Baseline & 98.18(99.99) & 98.00(99.99) & 99.28(100.00) & 99.65(99.99) & 100.00(100.00) & 99.02(100.00) \\
         & SFT & 98.50(99.99) & 100.00(100.00) & 100.00(100.00) & 99.93(100.00) & 99.92(99.99) & 99.67(100.00) \\
         & DPO & 98.55(100.00) & 99.50(100.00) & 100.00(100.00) & 99.67(99.99) & 100.00(100.00) & 99.54(100.00) \\
         & ORPO & 98.11(99.99) & 100.00(100.00) & 99.76(100.00) & 99.59(100.00) & 99.92(100.00) & 99.48(100.00) \\
         & TLPO(ours) & 98.83(100.00) & 100.00(100.00) & 100.00(100.00) & 99.83(100.00) & 100.00(100.00) & 99.73(100.00) \\ 

    \midrule 
    \multirow{5}{*}{\shortstack{\textbf{ar}}}  
         & Baseline & 91.67(99.96) & 94.67(99.97) & 95.96(99.92) & 96.10(99.97) & 96.44(99.91) & 94.97(99.94) \\
         & SFT & 97.37(99.98) & 99.67(99.99) & 99.76(99.87) & 99.61(99.99) & 99.17(99.97) & 99.12(99.96) \\
         & DPO & 96.42(99.98) & 99.00(99.99) & 99.52(99.99) & 98.23(99.98) & 98.51(99.94) & 98.34(99.98) \\
         & ORPO & 86.59(99.92) & 90.33(99.90) & 95.72(99.96) & 93.36(99.94) & 94.21(99.86) & 92.05(99.92) \\
         & TLPO(ours) & 98.18(99.99) & 99.00(99.99) & 99.52(100.00) & 99.45(100.00) & 99.01(99.97) & 99.03(99.99) \\

    \midrule 
    \multirow{5}{*}{\shortstack{\textbf{ko}}}
         & Baseline & 94.36(99.96) & 95.00(99.94) & 96.89(99.96) & 94.68(99.95) & 96.94(99.81) & 95.57(99.92) \\
         & SFT & 96.89(99.85) & 100.00(100.00) & 99.51(99.99) & 99.52(99.98) & 99.83(99.98) & 99.15(99.96) \\
         & DPO & 96.62(99.98) & 98.00(99.98) & 99.28(99.99) & 98.07(99.98) & 98.43(99.94) & 98.08(99.97) \\
         & ORPO & 88.06(99.90) & 93.00(99.91) & 95.20(99.92) & 89.63(99.88) & 95.53(99.74) & 92.29(99.87) \\
         & TLPO(ours) & 97.69(99.99) & 99.00(99.99) & 99.28(99.98) & 99.23(99.99) & 99.34(99.94) & 98.91(99.98) \\

    \midrule 
    \multirow{5}{*}{\shortstack{\textbf{ja}}}  
         & Baseline & 98.57(100.00) & 95.00(99.98) & 98.81(99.98) & 99.58(100.00) & 99.67(99.99) & 98.32(99.99) \\
         & SFT & 97.90(99.99) & 100.00(100.00) & 99.52(99.99) & 99.93(100.00) & 100.00(100.00) & 99.47(100.00) \\
         & DPO & 98.90(99.99) & 98.99(99.99) & 99.49(99.99) & 98.99(99.95) & 99.71(99.98) & 99.22(99.98) \\
         & ORPO & 98.55(99.98) & 100.00(100.00) & 98.52(99.92) & 99.09(99.96) & 99.50(99.97) & 99.13(99.97) \\
         & TLPO(ours) & 100.00(100.00) & 100.00(100.00) & 99.76(100.00) & 99.83(100.00) & 100.00(100.00) & 99.92(100.00) \\

    \midrule 
    \multirow{5}{*}{\shortstack{\textbf{avg.}}}  
         & Baseline & 95.69(99.98) & 95.67(99.97) & 97.74(99.96) & 97.50(99.98) & 98.26(99.93) & 96.97(99.96) \\
         & SFT & 97.66(99.95) & 99.92(100.00) & 99.70(99.96) & 99.75(99.99) & 99.73(99.99) & 99.35(99.98) \\
         & DPO & 97.62(99.99) & 98.87(99.99) & 99.58(99.99) & 98.74(99.98) & 99.16(99.96) & 98.79(99.98) \\
         & ORPO & 92.83(99.95) & 95.83(99.95) & 97.30(99.95) & 95.42(99.95) & 97.29(99.89) & 95.73(99.94) \\
         & TLPO(ours) & 98.68(99.99) & 99.50(100.00) & 99.64(99.99) & 99.58(100.00) & 99.59(99.98) & 99.40(99.99) \\

    \bottomrule         
    \end{tabular}

    \caption{Average Response Pass Rate(RPR) and Word Pass Rate(WPR). Values are presented as RPR(WPR) in \%.}        
    \end{subtable}

    \par\bigskip

    \begin{subtable}{0.98\textwidth}
    \centering
    \begin{tabular}{c||c||cccccccccc|c}
    \toprule
         \multirow{2}{*}{\fontsize{7pt}{9pt}\selectfont Lang.} 
         & \multirow{2}{*}{\fontsize{7pt}{9pt}\selectfont Method} 
         & \fontsize{7pt}{9pt}\selectfont MIF
         & \fontsize{7pt}{9pt}\selectfont MIF
         & \fontsize{6.5pt}{9pt}\selectfont MMMLU
         & \fontsize{7pt}{9pt}\selectfont GPQA
         & \fontsize{6.5pt}{9pt}\selectfont GPQA-D
         & \fontsize{6.5pt}{9pt}\selectfont ARC-C
         & \fontsize{7pt}{9pt}\selectfont BBH
         & \fontsize{7pt}{9pt}\selectfont MATH
         & \fontsize{7pt}{9pt}\selectfont GSM8K
         & \fontsize{7pt}{9pt}\selectfont GSM8K
         & \multirow{2}{*}{\fontsize{7pt}{9pt}\selectfont Mean} \\

         &
         & \fontsize{7pt}{9pt}\selectfont (en)       
         & \fontsize{7pt}{9pt}\selectfont (target) 
         & \fontsize{7pt}{9pt}\selectfont (target) 
         & \fontsize{7pt}{9pt}\selectfont (en)  
         & \fontsize{7pt}{9pt}\selectfont (en)
         & \fontsize{7pt}{9pt}\selectfont (en)  
         & \fontsize{7pt}{9pt}\selectfont (en)      
         & \fontsize{7pt}{9pt}\selectfont (en)  
         & \fontsize{7pt}{9pt}\selectfont (en)  
         & \fontsize{7pt}{9pt}\selectfont (cross) 
         &  \\

    \midrule 
    
    \midrule     
    \multirow{5}{*}{\fontsize{7pt}{9pt}\selectfont \shortstack{\textbf{zh}}}     
         & Baseline & 82.07 & 67.28 & 73.69 & 45.76 & 46.46 & 90.44 & 40.35 & 74.74 & 78.33 & 92.72 & 69.18 \\
         & SFT & 55.27 & 41.22 & 62.11 & 33.93 & 32.32 & 91.47 & 55.63 & 10.06 & 38.30 & 79.49 & 49.98 \\
         & DPO & 81.33 & 65.80 & 72.97 & 45.09 & 45.96 & 90.53 & 40.04 & 74.38 & 75.85 & 92.72 & 68.47 \\
         & ORPO & 81.89 & 64.51 & 73.05 & 43.75 & 44.95 & 91.13 & 41.91 & 75.60 & 63.94 & 92.56 & 67.33 \\
         & \fontsize{7pt}{9pt}\selectfont{TLPO(ours)} & 82.62 & 67.47 & 72.38 & 45.54 & 51.01 & 90.78 & 46.97 & 74.30 & 82.38 & 91.56 & 70.50 \\

    \midrule     
    \multirow{5}{*}{\fontsize{7pt}{9pt}\selectfont \shortstack{\textbf{ar}}}     
         & Baseline & 82.26 & 67.65 & 64.09 & 45.76 & 46.46 & 90.44 & 40.35 & 74.74 & 78.33 & 90.90 & 68.10 \\
         & SFT & 78.37 & 54.53 & 53.30 & 37.72 & 38.38 & 91.30 & 59.68 & 73.44 & 79.40 & 74.44 & 64.06 \\
         & DPO & 81.33 & 66.73 & 64.04 & 40.85 & 48.48 & 90.44 & 40.16 & 74.64 & 77.75 & 90.82 & 67.52 \\
         & ORPO & 81.70 & 67.65 & 64.62 & 43.97 & 48.99 & 90.96 & 40.59 & 75.54 & 71.30 & 90.65 & 67.60 \\
         & \fontsize{7pt}{9pt}\selectfont{TLPO(ours)} & 82.99 & 67.28 & 63.67 & 44.42 & 48.48 & 90.61 & 41.76 & 74.80 & 84.12 & 90.16 & 68.83 \\

    \midrule     
    \multirow{5}{*}{\fontsize{7pt}{9pt}\selectfont \shortstack{\textbf{ko}}}     
         & Baseline & 82.07 & 61.18 & 66.02 & 45.76 & 46.46 & 90.44 & 40.35 & 74.74 & 78.33 & 91.98 & 67.73 \\
         & SFT & 76.34 & 50.83 & 55.65 & 33.93 & 32.83 & 91.38 & 57.99 & 72.60 & 79.24 & 80.40 & 63.12 \\
         & DPO & 81.15 & 59.70 & 65.68 & 43.08 & 51.52 & 90.44 & 39.75 & 75.04 & 78.16 & 91.56 & 67.61 \\
         & ORPO & 80.96 & 61.18 & 65.77 & 43.08 & 51.01 & 90.78 & 40.55 & 75.42 & 71.22 & 92.22 & 67.22 \\
         & \fontsize{7pt}{9pt}\selectfont{TLPO(ours)} & 83.18 & 60.63 & 65.57 & 43.30 & 49.49 & 90.53 & 41.38 & 73.88 & 80.73 & 90.16 & 67.88 \\

    \midrule     
    \multirow{5}{*}{\fontsize{7pt}{9pt}\selectfont \shortstack{\textbf{ja}}}     
         & Baseline & 82.26 & 62.11 & 67.68 & 45.76 & 46.46 & 90.44 & 40.35 & 74.74 & 78.33 & 89.99 & 67.81 \\
         & SFT & 58.41 & 39.00 & 56.45 & 29.69 & 29.29 & 91.30 & 51.68 & 10.98 & 32.51 & 72.95 & 47.23 \\
         & DPO & 34.01 & 2.22 & 45.66 & 27.23 & 27.27 & 90.61 & 31.73 & 7.94 & 26.55 & 51.20 & 34.44 \\
         & ORPO & 35.30 & 2.59 & 50.49 & 32.81 & 29.80 & 90.96 & 33.33 & 11.82 & 25.56 & 59.72 & 37.24 \\
         & \fontsize{7pt}{9pt}\selectfont{TLPO(ours)} & 79.85 & 55.45 & 68.04 & 43.97 & 47.98 & 90.61 & 44.99 & 74.52 & 83.71 & 91.65 & 68.08 \\

    \midrule     
    \multirow{5}{*}{\fontsize{7pt}{9pt}\selectfont \shortstack{\textbf{avg.}}}     
         & Baseline & 82.16 & 64.56 & 67.87 & 45.76 & 46.46 & 90.44 & 40.35 & 74.74 & 78.33 & 91.40 & 68.21 \\
         & SFT & 67.10 & 46.40 & 56.87 & 33.82 & 33.21 & 91.36 & 56.25 & 41.77 & 57.36 & 76.82 & 56.10 \\
         & DPO & 69.46 & 48.61 & 62.09 & 39.06 & 43.31 & 90.51 & 37.92 & 58.00 & 64.58 & 81.57 & 59.51 \\
         & ORPO & 69.96 & 48.98 & 63.48 & 40.90 & 43.69 & 90.95 & 39.10 & 59.59 & 58.00 & 83.79 & 59.85 \\
         & \fontsize{7pt}{9pt}\selectfont{TLPO(ours)} & 82.16 & 62.71 & 67.41 & 44.31 & 49.24 & 90.64 & 43.77 & 74.38 & 82.73 & 90.88 & 68.82 \\

    \bottomrule         
    \end{tabular}

    \caption{Average accuracy after fine-tuning.}        
    \end{subtable}

    \caption{
    Detailed RPR, WPR, and accuracy results for the Qwen3-8B model after fine-tuning, in a setting where English output is regarded as neutral.
    }
    
    \label{tab:main_result_qwen}
    \vspace{-10pt}    
\end{table*}


\begin{table*}[!ht]
    \fontsize{7.6pt}{9pt}\selectfont
    \centering
    
    \begin{subtable}{0.98\textwidth}
    \centering
    \begin{tabular}{c||c||ccccc|c}
    \toprule
    \multirow{2}{*}{Lang.} & \multirow{2}{*}{Method} & LCB & LCB & MIF & MMMLU & GSM8K  & \multirow{2}{*}{Mean} \\
    & & (cross-lingual) & (monolingual) & (target) & (target) & (cross)  & \\
    \midrule 
    
    \midrule     
    \multirow{5}{*}{\shortstack{\textbf{zh}}}  
         & Baseline & 98.74(99.99) & 100.00(100.00) & 99.76(100.00) & 99.68(99.95) & 100.00(100.00) & 99.64(99.99) \\
         & SFT & 99.49(100.00) & 100.00(100.00) & 100.00(100.00) & 99.84(99.97) & 100.00(100.00) & 99.87(99.99) \\
         & DPO & 98.70(99.99) & 100.00(100.00) & 100.00(100.00) & 99.69(99.96) & 100.00(100.00) & 99.68(99.99) \\
         & ORPO & 98.27(99.99) & 100.00(100.00) & 100.00(100.00) & 99.77(99.98) & 100.00(100.00) & 99.61(99.99) \\
         & TLPO(ours) & 98.90(100.00) & 100.00(100.00) & 100.00(100.00) & 99.88(99.99) & 100.00(100.00) & 99.75(100.00) \\

    \midrule 
    \multirow{5}{*}{\shortstack{\textbf{ar}}}  
         & Baseline & 89.66(99.93) & 96.00(99.97) & 98.07(99.99) & 97.61(99.62) & 97.93(99.94) & 95.85(99.89) \\
         & SFT & 96.03(99.97) & 99.67(100.00) & 100.00(100.00) & 99.25(99.85) & 99.22(99.97) & 98.83(99.96) \\
         & DPO & 97.44(98.13) & 95.67(98.23) & 98.80(97.87) & 97.12(94.62) & 99.75(99.85) & 97.75(97.74) \\
         & ORPO & 96.55(99.99) & 98.33(99.99) & 99.28(99.98) & 98.34(99.74) & 99.50(99.96) & 98.40(99.93) \\
         & TLPO(ours) & 92.96(99.96) & 99.33(100.00) & 100.00(100.00) & 99.03(99.82) & 100.00(100.00) & 98.26(99.95) \\

    \midrule 
    \multirow{5}{*}{\shortstack{\textbf{ko}}}
         & Baseline & 95.79(99.93) & 100.00(100.00) & 97.82(99.98) & 97.66(99.87) & 98.75(99.95) & 98.01(99.94) \\
         & SFT & 98.24(99.99) & 100.00(100.00) & 99.50(99.99) & 99.47(99.92) & 99.56(99.98) & 99.35(99.98) \\
         & DPO & 98.82(100.00) & 100.00(100.00) & 99.76(99.83) & 98.57(98.22) & 99.69(99.97) & 99.37(99.60) \\
         & ORPO & 97.24(98.33) & 100.00(100.00) & 98.56(99.23) & 98.21(99.06) & 99.36(99.46) & 98.68(99.21) \\
         & TLPO(ours) & 98.40(99.99) & 100.00(100.00) & 100.00(100.00) & 99.52(99.98) & 100.00(100.00) & 99.58(99.99) \\

    \midrule 
    \multirow{5}{*}{\shortstack{\textbf{ja}}}  
         & Baseline & 99.19(100.00) & 99.00(100.00) & 99.28(99.97) & 99.65(99.99) & 99.92(99.99) & 99.41(99.99) \\
         & SFT & 98.28(99.20) & 100.00(100.00) & 99.25(99.97) & 99.74(99.97) & 99.81(99.99) & 99.42(99.83) \\
         & DPO & 99.19(100.00) & 100.00(100.00) & 99.76(99.97) & 99.61(99.98) & 99.92(99.98) & 99.70(99.99) \\
         & ORPO & 99.25(100.00) & 100.00(100.00) & 99.76(99.98) & 99.77(100.00) & 100.00(100.00) & 99.76(100.00) \\
         & TLPO(ours) & 99.57(100.00) & 100.00(100.00) & 99.76(99.99) & 99.85(99.99) & 100.00(100.00) & 99.84(100.00) \\

    \midrule 
    \multirow{5}{*}{\shortstack{\textbf{avg.}}}  
         & Baseline & 95.84(99.96) & 98.75(99.99) & 98.73(99.98) & 98.65(99.86) & 99.15(99.97) & 98.22(99.95) \\
         & SFT & 98.01(99.79) & 99.92(100.00) & 99.69(99.99) & 99.58(99.93) & 99.65(99.99) & 99.37(99.94) \\
         & DPO & 98.54(99.53) & 98.92(99.56) & 99.58(99.42) & 98.75(98.19) & 99.84(99.95) & 99.12(99.33) \\
         & ORPO & 97.83(99.58) & 99.58(100.00) & 99.40(99.80) & 99.02(99.69) & 99.72(99.85) & 99.11(99.78) \\
         & TLPO(ours) & 97.45(99.99) & 99.83(100.00) & 99.94(100.00) & 99.57(99.94) & 100.00(100.00) & 99.36(99.99) \\

    \bottomrule         
    \end{tabular}

    \caption{Average Response Pass Rate(RPR) and Word Pass Rate(WPR). Values are presented as RPR(WPR) in \%.}        
    \end{subtable}

    \par\bigskip

    \begin{subtable}{0.98\textwidth}
    \centering
    \begin{tabular}{c||c||cccccccccc|c}
    \toprule
         \multirow{2}{*}{\fontsize{7pt}{9pt}\selectfont Lang.} 
         & \multirow{2}{*}{\fontsize{7pt}{9pt}\selectfont Method} 
         & \fontsize{7pt}{9pt}\selectfont MIF
         & \fontsize{7pt}{9pt}\selectfont MIF
         & \fontsize{6.5pt}{9pt}\selectfont MMMLU
         & \fontsize{7pt}{9pt}\selectfont GPQA
         & \fontsize{6.5pt}{9pt}\selectfont GPQA-D
         & \fontsize{6.5pt}{9pt}\selectfont ARC-C
         & \fontsize{7pt}{9pt}\selectfont BBH
         & \fontsize{7pt}{9pt}\selectfont MATH
         & \fontsize{7pt}{9pt}\selectfont GSM8K
         & \fontsize{7pt}{9pt}\selectfont GSM8K
         & \multirow{2}{*}{\fontsize{7pt}{9pt}\selectfont Mean} \\

         &
         & \fontsize{7pt}{9pt}\selectfont (en)       
         & \fontsize{7pt}{9pt}\selectfont (target) 
         & \fontsize{7pt}{9pt}\selectfont (target) 
         & \fontsize{7pt}{9pt}\selectfont (en)  
         & \fontsize{7pt}{9pt}\selectfont (en)
         & \fontsize{7pt}{9pt}\selectfont (en)  
         & \fontsize{7pt}{9pt}\selectfont (en)      
         & \fontsize{7pt}{9pt}\selectfont (en)  
         & \fontsize{7pt}{9pt}\selectfont (en)  
         & \fontsize{7pt}{9pt}\selectfont (cross) 
         &  \\

    \midrule 
    
    \midrule     
    \multirow{5}{*}{\fontsize{7pt}{9pt}\selectfont \shortstack{\textbf{zh}}}     
         & Baseline & 53.60 & 44.36 & 50.28 & 30.80 & 26.77 & 80.97 & 52.57 & 52.38 & 79.40 & 85.28 & 55.64 \\
         & SFT & 49.35 & 39.37 & 43.81 & 27.68 & 22.22 & 81.48 & 55.46 & 40.60 & 78.41 & 75.19 & 51.36 \\
         & DPO & 54.53 & 44.55 & 49.90 & 29.69 & 27.78 & 81.06 & 52.02 & 52.52 & 79.74 & 84.37 & 55.61 \\
         & ORPO & 52.87 & 41.40 & 46.75 & 25.89 & 24.75 & 81.66 & 52.22 & 52.60 & 78.58 & 85.77 & 54.25 \\
         & \fontsize{7pt}{9pt}\selectfont{TLPO(ours)} & 50.83 & 42.70 & 48.14 & 25.67 & 28.79 & 81.14 & 51.71 & 51.78 & 78.91 & 85.19 & 54.49 \\

    \midrule     
    \multirow{5}{*}{\fontsize{7pt}{9pt}\selectfont \shortstack{\textbf{ar}}}     
         & Baseline & 53.60 & 36.23 & 43.16 & 30.80 & 26.77 & 80.97 & 52.57 & 52.38 & 79.40 & 76.26 & 53.22 \\
         & SFT & 52.68 & 32.16 & 37.32 & 28.79 & 29.29 & 81.91 & 55.41 & 40.94 & 78.49 & 68.57 & 50.56 \\
         & DPO & 56.01 & 34.94 & 40.95 & 29.69 & 25.76 & 81.40 & 52.86 & 51.64 & 78.33 & 78.25 & 52.98 \\
         & ORPO & 51.76 & 35.49 & 42.96 & 30.13 & 33.84 & 81.66 & 50.81 & 53.42 & 79.16 & 70.89 & 53.01 \\
         & \fontsize{7pt}{9pt}\selectfont{TLPO(ours)} & 54.34 & 36.60 & 39.86 & 29.69 & 29.29 & 81.06 & 51.97 & 51.32 & 79.07 & 80.48 & 53.37 \\

    \midrule     
    \multirow{5}{*}{\fontsize{7pt}{9pt}\selectfont \shortstack{\textbf{ko}}}     
         & Baseline & 53.42 & 39.19 & 47.54 & 30.80 & 26.77 & 80.97 & 52.57 & 52.38 & 79.40 & 80.07 & 54.31 \\
         & SFT & 51.39 & 35.12 & 41.61 & 26.12 & 30.81 & 81.57 & 54.49 & 40.36 & 78.66 & 73.61 & 51.37 \\
         & DPO & 55.64 & 39.74 & 46.00 & 27.01 & 27.78 & 80.89 & 51.80 & 51.22 & 79.82 & 77.58 & 53.75 \\
         & ORPO & 52.50 & 35.86 & 45.34 & 29.69 & 22.73 & 81.06 & 49.79 & 52.02 & 77.83 & 76.76 & 52.36 \\
         & \fontsize{7pt}{9pt}\selectfont{TLPO(ours)} & 52.87 & 37.15 & 47.57 & 31.70 & 31.31 & 81.48 & 52.03 & 53.24 & 78.16 & 77.34 & 54.29 \\

    \midrule     
    \multirow{5}{*}{\fontsize{7pt}{9pt}\selectfont \shortstack{\textbf{ja}}}     
         & Baseline & 53.60 & 39.37 & 49.24 & 30.80 & 26.77 & 80.97 & 52.57 & 52.38 & 79.40 & 84.20 & 54.93 \\
         & SFT & 52.87 & 31.98 & 42.24 & 25.22 & 25.25 & 81.91 & 55.14 & 42.34 & 79.24 & 72.54 & 50.87 \\
         & DPO & 54.34 & 41.04 & 47.25 & 31.92 & 27.27 & 81.06 & 50.62 & 51.42 & 78.66 & 84.20 & 54.78 \\
         & ORPO & 53.42 & 34.01 & 48.62 & 27.23 & 26.77 & 81.23 & 51.04 & 52.62 & 78.74 & 80.48 & 53.42 \\
         & \fontsize{7pt}{9pt}\selectfont{TLPO(ours)} & 52.31 & 37.34 & 49.24 & 27.23 & 29.80 & 81.14 & 52.23 & 51.94 & 78.66 & 82.13 & 54.20 \\

    \midrule     
    \multirow{5}{*}{\fontsize{7pt}{9pt}\selectfont \shortstack{\textbf{avg.}}}     
         & Baseline & 53.56 & 39.79 & 47.55 & 30.80 & 26.77 & 80.97 & 52.57 & 52.38 & 79.40 & 81.45 & 54.52 \\
         & SFT & 51.57 & 34.66 & 41.25 & 26.95 & 26.89 & 81.72 & 55.13 & 41.06 & 78.70 & 72.48 & 51.04 \\
         & DPO & 55.13 & 40.06 & 46.03 & 29.58 & 27.15 & 81.10 & 51.83 & 51.70 & 79.14 & 81.10 & 54.28 \\
         & ORPO & 52.64 & 36.69 & 45.92 & 28.24 & 27.02 & 81.40 & 50.96 & 52.66 & 78.58 & 78.47 & 53.26 \\
         & \fontsize{7pt}{9pt}\selectfont{TLPO(ours)} & 52.59 & 38.45 & 46.20 & 28.57 & 29.80 & 81.21 & 51.99 & 52.07 & 78.70 & 81.29 & 54.09 \\

    \bottomrule         
    \end{tabular}

    \caption{Average accuracy after fine-tuning.}        
    \end{subtable}

    \caption{
    Detailed RPR, WPR, and accuracy results for the Ministral-8B model after fine-tuning, in a setting where English output is regarded as neutral.
    }
    
    \label{tab:main_result_ministral}
    \vspace{-10pt}    
\end{table*}


\begin{table*}[!ht]
    \fontsize{7.6pt}{9pt}\selectfont
    \centering
    
    \begin{subtable}{0.98\textwidth}
    \centering
    \begin{tabular}{c||c||ccccc|c}
    \toprule
    \multirow{2}{*}{Lang.} & \multirow{2}{*}{Method} & LCB & LCB & MIF & MMMLU & GSM8K  & \multirow{2}{*}{Mean} \\
    & & (cross-lingual) & (monolingual) & (target) & (target) & (cross)  & \\
    \midrule 
    
    \midrule     
    \multirow{5}{*}{\shortstack{\textbf{zh}}}  
         & Baseline & 88.73(99.93) & 95.00(99.95) & 97.83(99.98) & 99.13(99.49) & 99.21(99.85) & 95.98(99.84) \\
         & SFT & 97.36(99.96) & 100.00(100.00) & 99.76(100.00) & 98.83(98.67) & 99.91(100.00) & 99.17(99.72) \\
         & DPO & 91.30(99.95) & 94.50(99.95) & 97.36(99.98) & 99.22(99.61) & 99.47(99.90) & 96.37(99.88) \\
         & ORPO & 89.01(99.87) & 99.00(99.99) & 97.12(99.97) & 98.86(99.05) & 99.75(99.92) & 96.75(99.76) \\
         & TLPO(ours) & 95.96(99.97) & 99.50(100.00) & 98.80(99.99) & 99.53(99.94) & 100.00(100.00) & 98.76(99.98) \\

    \midrule 
    \multirow{5}{*}{\shortstack{\textbf{ar}}}  
         & Baseline & 83.81(98.64) & 95.00(99.95) & 96.90(99.98) & 98.23(99.81) & 99.42(99.97) & 94.67(99.67) \\
         & SFT & 95.26(99.72) & 99.67(100.00) & 99.76(100.00) & 96.60(98.35) & 100.00(100.00) & 98.26(99.61) \\
         & DPO & 82.73(98.44) & 97.00(99.98) & 98.33(99.99) & 98.20(99.80) & 99.50(99.97) & 95.15(99.64) \\
         & ORPO & 86.59(99.14) & 97.33(99.84) & 98.57(99.99) & 98.33(99.85) & 99.50(99.99) & 96.07(99.76) \\
         & TLPO(ours) & 95.40(99.79) & 98.67(99.99) & 99.05(99.99) & 98.84(99.90) & 99.42(99.99) & 98.28(99.93) \\

    \midrule 
    \multirow{5}{*}{\shortstack{\textbf{ko}}}
         & Baseline & 96.74(99.96) & 98.00(99.98) & 96.90(99.95) & 98.30(99.85) & 93.65(99.87) & 96.72(99.92) \\
         & SFT & 97.04(99.92) & 100.00(100.00) & 99.75(100.00) & 98.14(99.75) & 99.91(100.00) & 98.97(99.93) \\
         & DPO & 96.38(99.96) & 100.00(100.00) & 97.37(99.94) & 98.33(99.86) & 93.78(99.88) & 97.17(99.93) \\
         & ORPO & 97.07(99.98) & 99.00(100.00) & 96.90(99.93) & 98.30(99.85) & 99.92(100.00) & 98.24(99.95) \\
         & TLPO(ours) & 97.43(99.97) & 100.00(100.00) & 98.09(99.95) & 98.48(99.88) & 97.10(99.95) & 98.22(99.95) \\

    \midrule 
    \multirow{5}{*}{\shortstack{\textbf{ja}}}  
         & Baseline & 91.67(99.95) & 92.00(99.91) & 94.98(99.89) & 99.14(99.92) & 97.23(99.84) & 95.00(99.90) \\
         & SFT & 97.10(99.96) & 100.00(100.00) & 98.01(99.53) & 95.55(99.55) & 100.00(100.00) & 98.13(99.81) \\
         & DPO & 94.58(99.84) & 93.00(99.83) & 95.22(99.85) & 99.10(99.75) & 97.73(99.97) & 95.93(99.85) \\
         & ORPO & 97.45(99.99) & 95.00(99.96) & 97.14(99.92) & 99.17(99.88) & 99.83(100.00) & 97.72(99.95) \\
         & TLPO(ours) & 97.46(99.99) & 100.00(100.00) & 98.57(99.99) & 99.38(99.90) & 99.33(99.99) & 98.95(99.98) \\

    \midrule 
    \multirow{5}{*}{\shortstack{\textbf{avg.}}}  
         & Baseline & 90.24(99.62) & 95.00(99.95) & 96.65(99.95) & 98.70(99.77) & 97.38(99.88) & 95.59(99.83) \\
         & SFT & 96.69(99.89) & 99.92(100.00) & 99.32(99.88) & 97.28(99.08) & 99.95(100.00) & 98.63(99.77) \\
         & DPO & 91.25(99.55) & 96.12(99.94) & 97.07(99.94) & 98.71(99.75) & 97.62(99.93) & 96.16(99.82) \\
         & ORPO & 92.53(99.74) & 97.58(99.95) & 97.43(99.95) & 98.67(99.66) & 99.75(99.97) & 97.19(99.86) \\
         & TLPO(ours) & 96.56(99.93) & 99.54(100.00) & 98.63(99.98) & 99.06(99.90) & 98.96(99.98) & 98.55(99.96) \\

    \bottomrule         
    \end{tabular}

    \caption{Average Response Pass Rate(RPR) and Word Pass Rate(WPR). Values are presented as RPR(WPR) in \%.}        
    \end{subtable}

    \par\bigskip

    \begin{subtable}{0.98\textwidth}
    \centering
    \begin{tabular}{c||c||cccccccccc|c}
    \toprule
         \multirow{2}{*}{\fontsize{7pt}{9pt}\selectfont Lang.} 
         & \multirow{2}{*}{\fontsize{7pt}{9pt}\selectfont Method} 
         & \fontsize{7pt}{9pt}\selectfont MIF
         & \fontsize{7pt}{9pt}\selectfont MIF
         & \fontsize{6.5pt}{9pt}\selectfont MMMLU
         & \fontsize{7pt}{9pt}\selectfont GPQA
         & \fontsize{6.5pt}{9pt}\selectfont GPQA-D
         & \fontsize{6.5pt}{9pt}\selectfont ARC-C
         & \fontsize{7pt}{9pt}\selectfont BBH
         & \fontsize{7pt}{9pt}\selectfont MATH
         & \fontsize{7pt}{9pt}\selectfont GSM8K
         & \fontsize{7pt}{9pt}\selectfont GSM8K
         & \multirow{2}{*}{\fontsize{7pt}{9pt}\selectfont Mean} \\

         &
         & \fontsize{7pt}{9pt}\selectfont (en)       
         & \fontsize{7pt}{9pt}\selectfont (target) 
         & \fontsize{7pt}{9pt}\selectfont (target) 
         & \fontsize{7pt}{9pt}\selectfont (en)  
         & \fontsize{7pt}{9pt}\selectfont (en)
         & \fontsize{7pt}{9pt}\selectfont (en)  
         & \fontsize{7pt}{9pt}\selectfont (en)      
         & \fontsize{7pt}{9pt}\selectfont (en)  
         & \fontsize{7pt}{9pt}\selectfont (en)  
         & \fontsize{7pt}{9pt}\selectfont (cross) 
         &  \\

    \midrule 
    
    \midrule     
    \multirow{5}{*}{\fontsize{7pt}{9pt}\selectfont \shortstack{\textbf{zh}}}     
         & Baseline & 68.39 & 49.35 & 55.50 & 31.70 & 30.81 & 75.09 & 55.86 & 21.38 & 77.25 & 81.72 & 54.71 \\
         & SFT & 53.42 & 36.23 & 52.14 & 23.21 & 23.23 & 74.06 & 57.53 & 41.44 & 23.57 & 23.49 & 40.83 \\
         & DPO & 69.87 & 50.83 & 56.03 & 29.24 & 29.29 & 75.09 & 55.98 & 20.90 & 76.76 & 82.55 & 54.65 \\
         & ORPO & 68.76 & 50.65 & 54.64 & 27.90 & 28.28 & 76.02 & 55.83 & 20.18 & 77.17 & 77.75 & 53.72 \\
         & \fontsize{7pt}{9pt}\selectfont{TLPO(ours)} & 70.43 & 51.76 & 55.77 & 28.79 & 29.80 & 74.40 & 56.55 & 15.34 & 78.74 & 81.89 & 54.35 \\

    \midrule     
    \multirow{5}{*}{\fontsize{7pt}{9pt}\selectfont \shortstack{\textbf{ar}}}     
         & Baseline & 67.65 & 60.26 & 53.79 & 31.70 & 30.81 & 75.09 & 55.86 & 21.38 & 77.25 & 77.58 & 55.14 \\
         & SFT & 56.38 & 37.15 & 48.95 & 27.68 & 25.25 & 74.57 & 57.15 & 38.22 & 25.89 & 46.82 & 43.81 \\
         & DPO & 69.32 & 61.18 & 53.43 & 28.57 & 29.80 & 75.09 & 55.80 & 21.28 & 77.09 & 76.92 & 54.85 \\
         & ORPO & 69.32 & 58.96 & 52.83 & 25.67 & 24.75 & 75.77 & 56.29 & 21.02 & 77.09 & 78.25 & 53.99 \\
         & \fontsize{7pt}{9pt}\selectfont{TLPO(ours)} & 68.76 & 58.60 & 53.30 & 27.23 & 28.79 & 74.66 & 55.41 & 18.06 & 76.18 & 77.83 & 53.88 \\

    \midrule     
    \multirow{5}{*}{\fontsize{7pt}{9pt}\selectfont \shortstack{\textbf{ko}}}     
         & Baseline & 68.58 & 50.46 & 53.45 & 31.70 & 30.30 & 75.09 & 55.86 & 20.82 & 77.25 & 72.54 & 53.61 \\
         & SFT & 52.50 & 31.24 & 49.95 & 24.33 & 25.76 & 74.23 & 57.06 & 38.16 & 24.48 & 26.63 & 40.43 \\
         & DPO & 69.50 & 49.72 & 53.52 & 29.69 & 29.29 & 75.09 & 55.89 & 20.68 & 77.17 & 71.88 & 53.24 \\
         & ORPO & 68.21 & 47.13 & 53.60 & 27.68 & 31.31 & 75.34 & 55.35 & 13.16 & 77.75 & 57.24 & 50.68 \\
         & \fontsize{7pt}{9pt}\selectfont{TLPO(ours)} & 70.79 & 49.54 & 53.44 & 25.89 & 28.79 & 74.57 & 56.12 & 19.58 & 77.09 & 72.29 & 52.81 \\

    \midrule     
    \multirow{5}{*}{\fontsize{7pt}{9pt}\selectfont \shortstack{\textbf{ja}}}     
         & Baseline & 68.21 & 50.09 & 54.66 & 31.70 & 30.81 & 75.09 & 55.86 & 21.38 & 77.25 & 73.86 & 53.89 \\
         & SFT & 51.20 & 31.42 & 51.03 & 22.10 & 27.78 & 74.15 & 56.18 & 37.84 & 24.65 & 45.82 & 42.22 \\
         & DPO & 71.53 & 49.91 & 54.68 & 26.79 & 26.77 & 75.09 & 55.95 & 20.62 & 77.50 & 73.95 & 53.28 \\
         & ORPO & 69.50 & 47.32 & 54.13 & 26.79 & 29.80 & 75.68 & 56.63 & 15.66 & 77.34 & 75.68 & 52.85 \\
         & \fontsize{7pt}{9pt}\selectfont{TLPO(ours)} & 68.76 & 48.80 & 54.82 & 27.46 & 26.26 & 74.91 & 55.15 & 17.78 & 77.34 & 76.59 & 52.79 \\

    \midrule     
    \multirow{5}{*}{\fontsize{7pt}{9pt}\selectfont \shortstack{\textbf{avg.}}}     
         & Baseline & 68.21 & 52.54 & 54.35 & 31.70 & 30.68 & 75.09 & 55.86 & 21.24 & 77.25 & 76.43 & 54.33 \\
         & SFT & 53.37 & 34.01 & 50.52 & 24.33 & 25.51 & 74.25 & 56.98 & 38.91 & 24.65 & 35.69 & 41.82 \\
         & DPO & 70.06 & 52.91 & 54.42 & 28.57 & 28.79 & 75.09 & 55.91 & 20.87 & 77.13 & 76.32 & 54.01 \\
         & ORPO & 68.95 & 51.02 & 53.80 & 27.01 & 28.53 & 75.70 & 56.02 & 17.50 & 77.34 & 72.23 & 52.81 \\
         & \fontsize{7pt}{9pt}\selectfont{TLPO(ours)} & 69.69 & 52.17 & 54.33 & 27.34 & 28.41 & 74.64 & 55.81 & 17.69 & 77.34 & 77.15 & 53.46 \\

    \bottomrule         
    \end{tabular}

    \caption{Average accuracy after fine-tuning.}        
    \end{subtable}

    \caption{
    Detailed RPR, WPR, and accuracy results for the Gemma3-4B-it model after fine-tuning, in a setting where English output is regarded as neutral.
    }
    
    \label{tab:main_result_gemma}
    \vspace{-10pt}    
\end{table*}


\begin{table*}[!ht]
    \fontsize{7.6pt}{9pt}\selectfont
    \centering
    
    \begin{subtable}{0.98\textwidth}
    \centering
    \begin{tabular}{c||c||ccccc|c}
    \toprule
    \multirow{2}{*}{Lang.} & \multirow{2}{*}{Method} & LCB & LCB & MIF & MMMLU & GSM8K  & \multirow{2}{*}{Mean} \\
    & & (cross-lingual) & (monolingual) & (target) & (target) & (cross)  & \\
    \midrule 
    
    \midrule     
    \multirow{5}{*}{\shortstack{\textbf{zh}}}  
         & Baseline & 96.15(99.98) & 98.25(99.98) & 99.22(99.99) & 99.55(99.86) & 99.78(99.96) & 98.59(99.96) \\
         & SFT & 98.42(99.99) & 100.00(100.00) & 99.88(100.00) & 99.63(99.66) & 99.94(100.00) & 99.57(99.93) \\
         & DPO & 96.70(99.78) & 98.50(99.99) & 99.22(99.99) & 99.58(99.86) & 99.85(99.93) & 98.77(99.91) \\
         & ORPO & 95.57(99.95) & 99.38(99.99) & 99.04(99.99) & 99.48(99.75) & 99.88(99.98) & 98.67(99.93) \\
         & TLPO(ours) & 98.24(99.99) & 99.88(100.00) & 99.70(100.00) & 99.77(99.98) & 100.00(100.00) & 99.52(99.99) \\

    \midrule 
    \multirow{5}{*}{\shortstack{\textbf{ar}}}  
         & Baseline & 87.84(99.61) & 95.00(99.97) & 96.60(99.96) & 97.16(99.84) & 97.08(99.92) & 94.74(99.86) \\
         & SFT & 96.22(99.91) & 99.58(100.00) & 99.88(99.97) & 98.84(99.54) & 99.35(99.98) & 98.78(99.88) \\
         & DPO & 93.13(98.89) & 97.75(99.55) & 98.99(99.43) & 98.18(98.45) & 99.01(99.48) & 97.41(99.16) \\
         & ORPO & 90.11(99.75) & 95.42(99.93) & 98.04(99.98) & 96.95(99.88) & 97.39(99.90) & 95.58(99.89) \\
         & TLPO(ours) & 96.09(99.93) & 99.25(100.00) & 99.64(100.00) & 99.28(99.93) & 99.55(99.99) & 98.76(99.97) \\

    \midrule 
    \multirow{5}{*}{\shortstack{\textbf{ko}}}
         & Baseline & 93.66(99.91) & 94.75(99.94) & 96.05(99.96) & 96.38(99.90) & 95.81(99.83) & 95.33(99.91) \\
         & SFT & 97.44(99.94) & 100.00(100.00) & 99.32(99.99) & 99.19(99.91) & 99.66(99.97) & 99.12(99.96) \\
         & DPO & 97.34(99.98) & 99.50(100.00) & 98.98(99.94) & 98.64(99.51) & 97.89(99.95) & 98.47(99.87) \\
         & ORPO & 93.37(99.50) & 96.50(99.96) & 96.30(99.74) & 95.54(99.68) & 97.40(99.73) & 95.82(99.72) \\
         & TLPO(ours) & 97.62(99.96) & 99.75(100.00) & 99.10(99.98) & 99.17(99.96) & 98.92(99.96) & 98.91(99.97) \\

    \midrule 
    \multirow{5}{*}{\shortstack{\textbf{ja}}}  
         & Baseline & 96.98(99.98) & 96.50(99.97) & 98.21(99.96) & 99.53(99.98) & 99.20(99.96) & 98.08(99.97) \\
         & SFT & 97.57(99.78) & 100.00(100.00) & 99.07(99.87) & 98.79(99.88) & 99.93(100.00) & 99.07(99.91) \\
         & DPO & 98.07(99.96) & 97.75(99.95) & 98.50(99.95) & 99.37(99.91) & 99.32(99.97) & 98.60(99.95) \\
         & ORPO & 98.35(99.99) & 98.75(99.99) & 98.74(99.94) & 99.45(99.96) & 99.83(99.99) & 99.02(99.97) \\
         & TLPO(ours) & 98.78(100.00) & 100.00(100.00) & 99.40(99.98) & 99.73(99.97) & 99.83(100.00) & 99.55(99.99) \\

    \midrule 
    \multirow{5}{*}{\shortstack{\textbf{avg.}}}  
         & Baseline & 93.66(99.87) & 96.12(99.97) & 97.52(99.97) & 98.15(99.89) & 97.97(99.92) & 96.68(99.92) \\
         & SFT & 97.41(99.90) & 99.90(100.00) & 99.54(99.96) & 99.11(99.75) & 99.72(99.99) & 99.14(99.92) \\
         & DPO & 96.31(99.65) & 98.37(99.87) & 98.92(99.83) & 98.94(99.43) & 99.02(99.83) & 98.31(99.72) \\
         & ORPO & 94.35(99.80) & 97.51(99.97) & 98.03(99.91) & 97.85(99.82) & 98.63(99.90) & 97.27(99.88) \\
         & TLPO(ours) & 97.68(99.97) & 99.72(100.00) & 99.46(99.99) & 99.49(99.96) & 99.58(99.99) & 99.19(99.98) \\

    \bottomrule         
    \end{tabular}

    \caption{Average Response Pass Rate(RPR) and Word Pass Rate(WPR). Values are presented as RPR(WPR) in \%.}        
    \end{subtable}

    \par\bigskip

    \begin{subtable}{0.98\textwidth}
    \centering
    \begin{tabular}{c||c||cccccccccc|c}
    \toprule
         \multirow{2}{*}{\fontsize{7pt}{9pt}\selectfont Lang.} 
         & \multirow{2}{*}{\fontsize{7pt}{9pt}\selectfont Method} 
         & \fontsize{7pt}{9pt}\selectfont MIF
         & \fontsize{7pt}{9pt}\selectfont MIF
         & \fontsize{6.5pt}{9pt}\selectfont MMMLU
         & \fontsize{7pt}{9pt}\selectfont GPQA
         & \fontsize{6.5pt}{9pt}\selectfont GPQA-D
         & \fontsize{6.5pt}{9pt}\selectfont ARC-C
         & \fontsize{7pt}{9pt}\selectfont BBH
         & \fontsize{7pt}{9pt}\selectfont MATH
         & \fontsize{7pt}{9pt}\selectfont GSM8K
         & \fontsize{7pt}{9pt}\selectfont GSM8K
         & \multirow{2}{*}{\fontsize{7pt}{9pt}\selectfont Mean} \\

         &
         & \fontsize{7pt}{9pt}\selectfont (en)       
         & \fontsize{7pt}{9pt}\selectfont (target) 
         & \fontsize{7pt}{9pt}\selectfont (target) 
         & \fontsize{7pt}{9pt}\selectfont (en)  
         & \fontsize{7pt}{9pt}\selectfont (en)
         & \fontsize{7pt}{9pt}\selectfont (en)  
         & \fontsize{7pt}{9pt}\selectfont (en)      
         & \fontsize{7pt}{9pt}\selectfont (en)  
         & \fontsize{7pt}{9pt}\selectfont (en)  
         & \fontsize{7pt}{9pt}\selectfont (cross) 
         &  \\

    \midrule 
    
    \midrule     
    \multirow{5}{*}{\fontsize{7pt}{9pt}\selectfont \shortstack{\textbf{zh}}}     
         & Baseline & 69.69 & 53.14 & 58.70 & 33.93 & 32.58 & 82.57 & 50.01 & 49.43 & 78.56 & 84.84 & 59.34 \\
         & SFT & 57.35 & 41.22 & 49.99 & 27.62 & 25.13 & 82.79 & 56.38 & 33.86 & 54.26 & 62.53 & 49.11 \\
         & DPO & 70.29 & 52.91 & 58.15 & 32.59 & 32.32 & 82.59 & 49.77 & 49.14 & 77.81 & 84.88 & 59.05 \\
         & ORPO & 68.90 & 50.42 & 55.97 & 30.97 & 30.93 & 83.15 & 50.46 & 49.47 & 74.59 & 84.10 & 57.90 \\
         & \fontsize{7pt}{9pt}\selectfont{TLPO(ours)} & 69.36 & 54.21 & 58.00 & 30.97 & 34.09 & 82.34 & 52.00 & 47.74 & 79.98 & 84.76 & 59.35 \\

    \midrule     
    \multirow{5}{*}{\fontsize{7pt}{9pt}\selectfont \shortstack{\textbf{ar}}}     
         & Baseline & 69.45 & 52.54 & 52.06 & 33.65 & 33.08 & 82.53 & 50.12 & 49.57 & 78.47 & 79.90 & 58.14 \\
         & SFT & 64.70 & 42.88 & 43.99 & 30.75 & 29.55 & 82.94 & 57.65 & 49.22 & 65.28 & 63.90 & 53.09 \\
         & DPO & 70.10 & 51.80 & 50.51 & 32.14 & 33.33 & 82.62 & 50.22 & 49.10 & 78.23 & 79.76 & 57.78 \\
         & ORPO & 68.44 & 50.60 & 50.01 & 31.58 & 33.46 & 83.13 & 49.21 & 49.72 & 76.43 & 79.22 & 57.18 \\
         & \fontsize{7pt}{9pt}\selectfont{TLPO(ours)} & 69.18 & 50.14 & 49.86 & 31.92 & 34.22 & 82.51 & 50.31 & 47.82 & 79.67 & 78.41 & 57.40 \\

    \midrule     
    \multirow{5}{*}{\fontsize{7pt}{9pt}\selectfont \shortstack{\textbf{ko}}}     
         & Baseline & 69.73 & 47.92 & 54.04 & 33.76 & 32.20 & 82.55 & 49.94 & 49.27 & 78.64 & 80.11 & 57.82 \\
         & SFT & 63.54 & 39.60 & 45.43 & 27.96 & 29.80 & 82.74 & 56.88 & 48.70 & 64.72 & 62.55 & 52.19 \\
         & DPO & 70.43 & 47.37 & 53.00 & 31.25 & 32.83 & 82.53 & 49.50 & 48.78 & 78.53 & 78.53 & 57.27 \\
         & ORPO & 68.99 & 44.04 & 49.84 & 31.70 & 32.32 & 82.83 & 48.99 & 47.18 & 76.41 & 75.54 & 55.78 \\
         & \fontsize{7pt}{9pt}\selectfont{TLPO(ours)} & 69.69 & 48.43 & 53.52 & 32.03 & 34.47 & 82.55 & 50.38 & 48.96 & 78.85 & 79.09 & 57.80 \\

    \midrule     
    \multirow{5}{*}{\fontsize{7pt}{9pt}\selectfont \shortstack{\textbf{ja}}}     
         & Baseline & 69.78 & 48.29 & 55.50 & 33.31 & 32.32 & 82.55 & 50.20 & 49.56 & 78.25 & 81.24 & 58.10 \\
         & SFT & 58.96 & 35.95 & 46.72 & 26.62 & 27.02 & 82.79 & 55.18 & 33.64 & 53.14 & 64.39 & 48.44 \\
         & DPO & 58.41 & 34.66 & 49.39 & 27.40 & 27.65 & 82.64 & 47.36 & 32.14 & 65.67 & 71.11 & 49.64 \\
         & ORPO & 57.67 & 30.22 & 50.51 & 27.85 & 27.53 & 82.87 & 48.24 & 32.55 & 65.16 & 73.51 & 49.61 \\
         & \fontsize{7pt}{9pt}\selectfont{TLPO(ours)} & 68.67 & 46.07 & 55.55 & 31.42 & 32.70 & 82.68 & 50.91 & 48.31 & 79.71 & 81.68 & 57.77 \\

    \midrule     
    \multirow{5}{*}{\fontsize{7pt}{9pt}\selectfont \shortstack{\textbf{avg.}}}     
         & Baseline & 69.66 & 50.47 & 55.07 & 33.66 & 32.54 & 82.55 & 50.07 & 49.46 & 78.48 & 81.52 & 58.35 \\
         & SFT & 61.14 & 39.91 & 46.54 & 28.24 & 27.87 & 82.81 & 56.52 & 41.35 & 59.35 & 63.34 & 50.71 \\
         & DPO & 67.31 & 46.68 & 52.76 & 30.85 & 31.53 & 82.59 & 49.21 & 44.79 & 75.06 & 78.57 & 55.94 \\
         & ORPO & 66.00 & 43.82 & 51.58 & 30.52 & 31.06 & 82.99 & 49.22 & 44.73 & 73.14 & 78.09 & 55.12 \\
         & \fontsize{7pt}{9pt}\selectfont{TLPO(ours)} & 69.22 & 49.71 & 54.24 & 31.58 & 33.87 & 82.52 & 50.90 & 48.21 & 79.55 & 80.99 & 58.08 \\

    \bottomrule         
    \end{tabular}

    \caption{Average accuracy after fine-tuning.}        
    \end{subtable}

    \caption{
    Detailed RPR, WPR, and accuracy averaged across the 4 models after fine-tuning, in a setting where English output is regarded as neutral.
    }
    
    \label{tab:main_result_avg}
    \vspace{-10pt}    
\end{table*}

\end{document}